# An Integrated Causal Inference Framework for Traffic Safety Modeling with Semantic Street-View Visual Features

Lishan Sun[a], Yujia Cheng[a], Pengfei Cui[a,*], Lei Han[b], Mohamed Abdel-Aty[b], Yunhan Zheng[c,d], Xingchen Zhang[a]

[a] *College of Metropolitan Transportation, Beijing University of Technology, Beijing 100124, China*
[b] *Department of Civil, Environmental & Construction Engineering, University of Central Florida, Orlando, FL 32816, United States*
[c] *College of Urban and Environmental Sciences, Peking University, Beijing 100871, PR China*
[d] *Laboratory for Earth Surface Processes of the Ministry of Education, Peking University, Beijing 100871, PR China*

***Email addresses of authors****:* lssun@bjut.edu.cn, cyj03200@emails.bjut.edu.cn*,* pengfeicui@bjut.edu.cn, lei.han@ucf.edu, M.aty@ucf.edu, yunhan@pku.edu.cn, xczhang@bjut.edu.cn
***Corresponding author****:* pengfeicui@bjut.edu.cn

**Conflict of Interests:** *The authors declare that they have no known competing financial interests or personal relationships that could have appeared to influence the work reported in this paper.*

**Manuscript (Without Author Details)**

**Abstract:** Macroscopic traffic safety modeling aims to identify critical risk factors for regional crashes, thereby informing targeted policy interventions for safety improvement. However, current approaches rely heavily on static sociodemographic and infrastructure metrics, frequently overlooking the impacts from drivers' visual perception of driving environment. Although visual environment features have been found to impact driving and traffic crashes, existing evidence remains largely observational, failing to establish the robust causality for traffic policy evaluation under complex spatial environment. To fill these gaps, we applied semantic segmentation on Google Street View imageries to extract visual environmental features and proposed a Double Machine Learning framework to quantify their causal effects on regional crashes. Meanwhile, we utilized SHAP values to characterize the nonlinear influence mechanisms of confounding variables in the models and applied causal forests to estimate conditional average treatment effects. Leveraging crash records from the Miami metropolitan area, Florida, and 220,000 street view images, evidence shows that greenery proportion exerts a significant and robust negative causal effect on traffic crashes (*Average Treatment Effect = −6.38, p = 0.005*). This protective effect exhibits spatial heterogeneity, being most pronounced in densely populated and socially vulnerable urban cores. While greenery significantly mitigates angle and rear-end crashes, its protective benefit for vulnerable road users (VRUs) remains limited. Our findings provide causal evidence for greening as a potential safety intervention, prioritizing hazardous visual environments while highlighting the need for distinct design optimizations to protect VRUs.

# 1 Introduction

Urban traffic crashes continue to impose substantial social, economic, and public health burdens worldwide, demanding persistent research and policy intervention (Choi and Ewing, 2021). Despite substantial investment in infrastructure and traffic management, the decline in crash frequency has plateaued. This indicates that macro-scale engineering and operational strategies alone have likely reached a limit in their ability to deliver further safety gains (Ehsani et al., 2023). Recently, growing evidence reveals that micro-scale visual environment plays a critical, yet often underappreciated, role in shaping road user behavior through roadside elements, visual complexity, and streetscape perceptual cues (Nicholls et al., 2024). With the rapid expansion of street view platforms and advances in computer vision, visual environment features from street view imagery are increasingly being integrated into traffic safety modeling to characterize micro-scale environmental determinants (Cai et al., 2022; Cui et al., 2025; Fan et al., 2023).

Understanding the role of micro-scale visual environments in traffic safety is of substantial theoretical and practical importance (Yue, 2025). Theoretically, visual perception directly influences drivers' situational awareness, speed choice, and risk perception, thereby shaping crash occurrence through behavioral pathways that are not fully captured by traditional traffic variables (e.g., road geometry) (Jia et al., 2024). Practically, micro-scale visual features are closely linked to urban design and streetscape configuration, which are often more flexible and economical to modify than large-scale infrastructure elements. For example, recent studies utilizing street view imagery have demonstrated that specific visual elements, such as the proportion of greenery, spatial enclosure, and sidewalk visibility, are quantifiable determinants that significantly shape pedestrians' perceived road safety (Hamim and Ukkusuri, 2024). Therefore, exploring the causal impacts of such visual environment features on crash outcomes provide a stronger evidence for targeted microscopic design strategies that complement traditional traffic engineering measures (Yue, 2025).

Recent studies have increasingly incorporated street view imagery derived features into traffic safety modeling, demonstrating their strong predictive power in explaining crash frequency and severity (Park and Lee, 2026; Ye et al., 2025). However, most existing studies remain fundamentally associational, relying on statistical regression or machine learning models optimized for prediction rather than causal inference (Liu et al., 2025). Two key limitations persist: First, visual environment features derived from street view imagery are typically high-dimensional and strongly correlated, and they are often entangled with multi confounders related to land use, socioeconomic context, and roadway design, making causal interpretation challenging. Second, existing analyses largely focus on average or global associations, providing limited insight into whether the impacts of visual environments differ across urban contexts, locations, or crash types (Alisan and Ozguven, 2024). These limitations highlight a need for more rigorous causal inference and a deeper investigation of heterogeneous treatment effects in traffic safety research. Therefore, this study moves beyond correlational analysis to investigate the causal effects of micro-scale visual environment features on traffic safety. We focus on answering the following research questions:

(1) How can the Average Treatment Effects (ATEs) of micro-level visual environment features on traffic safety be estimated in observational data characterized by high-dimensional confounders?

(2) Do key interventions yield spatially heterogeneous safety benefits, and under which visual environment and socioeconomic conditions are these effects most pronounced?

(3) How can causal estimates be translated into geographically localized evidence to support fairer and targeted urban safety interventions?

The overall research logic, bridging the theoretical motivation and challenges to the proposed causal inference framework and intended contributions, is visualized in Fig. 1. We first employ computer vision techniques to extract visual features from street view imagery, constructing a dataset of the visual environment indicators. Second, we adopt the Double Machine Learning (DML) framework to remove confounding bias via orthogonalization, thereby obtaining unbiased estimates of the ATEs of key visual features. Third, we apply a Causal Forest model for the identified key factors to move beyond average effects and investigate their conditional average treatment effects (CATEs), revealing the spatial heterogeneity of causal effects and their variation across community contexts. The contributions of this study are threefold:

(1) Propose a two-stage causal inference framework tailored for high-dimensional SVI data, which effectively mitigates confounding bias to accurately estimate the ATEs of micro-level visual features.

(2) Elucidate the spatial heterogeneity safety benefits of key visual environment features by integrating robust DML with Causal Forests, identifying the specific visual and socioeconomic contexts where interventions are most effective.

(3) Translate causal estimates into geographically localized evidence to support decision-making, providing a scientific basis for designing fairer and more efficient urban safety interventions under resource constraints.

The remainder of this paper is organized as follows: Section 2 reviews the literature on traffic safety modeling and visual environment features. Section 3 describes the study area, data sources, and variable construction. Section 4 details the causal inference methodology comprising two stages. Section 5 presents empirical results from both the robust average treatment effect screening and the heterogeneity analysis. Section 6 concludes with key findings, policy implications, and directions for future research.

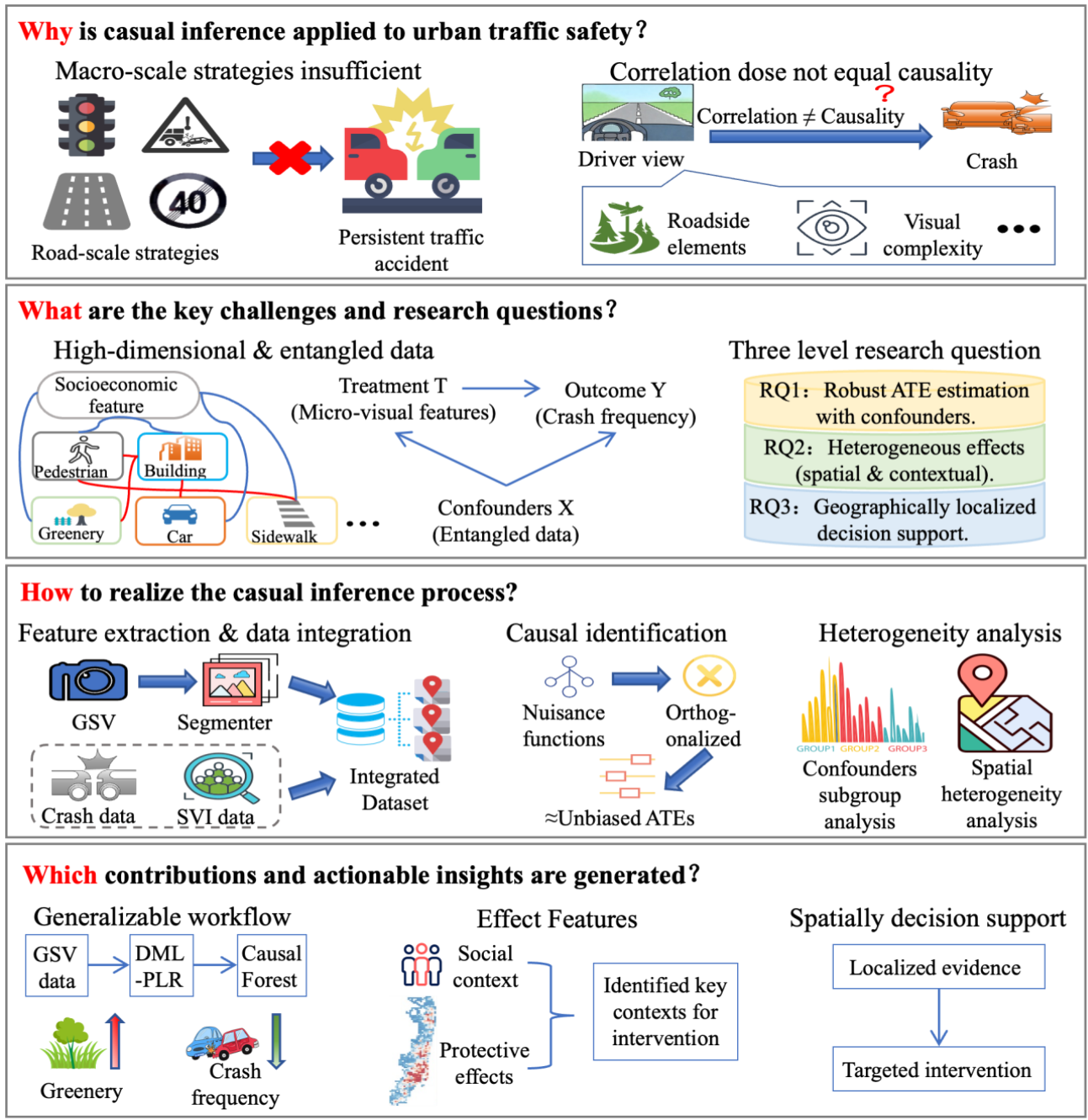

Fig. 1. Schematic overview of the research design addressing challenges in traffic safety causality.

## 2 Literature review

Building on the research motivation and analytical approach, this section reviews the progress of traffic safety research based on street view imagery and the growing adoption of causal inference methodologies. Table. 1 provides an overview of representative traffic safety studies, situating them across dimensions of visual environment representation, causal identification, and contextual effect heterogeneity.

2.1 Applications of street view imagery in traffic safety modeling

With advances in computer vision and urban sensing technologies, street view imagery has emerged as an important data source for characterizing the micro-scale visual environment. Unlike traditional measures derived from census data, GIS layers, or remote sensing, street view imagery captures urban spaces from a human centric perspective, enabling detailed observation of street level features such as road geometry, traffic facilities, and environmental context (Larkin et al., 2021; Xu et al., 2023). As a result, street view imagery has increasingly been adopted to explore how micro-scale environmental characteristics relate to traffic safety outcomes (Fan et al., 2025).

Early studies relied primarily on manual audits or perceptual assessments of street scenes, often using expert ratings or survey instruments to evaluate perceived safety, walkability, or visual quality (Fang et al., 2024; Hamim and Ukkusuri, 2024). Their reliance on human evaluation and experimental or survey data collection meant that they were not originally intended for extensive, continuous, or citywide applications. More recently, the application of deep learning and computer vision techniques has enabled the automated extraction of high-dimensional semantic features from large scale street view imagery (Y. Li et al., 2022). Using convolutional neural networks and semantic segmentation, researchers have quantified key visual elements such as greenery, visual complexity, and traffic participants (Pi et al., 2021). For instance, recent empirical evidence has revealed that the presence of trees is significantly correlated with reduced vehicle crash frequencies, likely by serving as vertical spatial definitions that enhance driver attentiveness (Yue, 2025). These quantified metrics have subsequently been linked to various safety outcomes, including traffic crashes, cycling safety, and driving behavior (Ye et al., 2024; Zhang et al., 2025). Empirical evidence consistently suggests that visual characteristics are significantly associated with safety outcomes, highlighting the potential of street view imagery to capture meaningful environmental signals.

2.2 Transitioning from observational associations to causal identification in traffic safety

Traffic safety research has traditionally focused on identifying associations between crash outcomes and observable covariates, rather than on estimating the causal effects of interventions or exposures (Zhang et al., 2021). In recent years, understanding the causal impacts of visual environment features on traffic safety outcomes has received increasing attention in transportation research, driven by the growing recognition that correlations alone cannot reveal the mechanisms underlying observed crash patterns (Ito et al., 2024). Unlike correlational analyses, causal inference seeks to estimate counterfactual outcomes to quantify the effects of hypothetical interventions. Currently, the application of causal inference methods is rapidly expanding, encompassing a variety of advanced strategies (Graham, 2025). For example, the instrumental variable approach addresses endogeneity by introducing exogenous variables (Wang et al., 2025); and double robust machine learning combines the advantages of regression and propensity score weighting to enhance estimation robustness (Li et al., 2024).

Although these methods have demonstrated significant value in addressing confounding bias, focusing solely on ATEs often masks the heterogeneity of environmental characteristic impacts on traffic safety (Kim, 2023). To capture effect heterogeneity within a causal framework, further advances in causal inference methods based on machine learning have focused on estimating CATEs. Among them, Causal Forests represent a prominent data-driven method that estimates CATEs by recursively partitioning the covariate space (Lechner and Mareckova, 2024). Unlike traditional spatial models, Causal Forests can flexibly accommodate nonlinear relationships and high-dimensional interactions without imposing strong parametric assumptions, enabling the discovery of heterogeneous treatment effects across the covariate space. Despite their strong performance in fields such as economics, the application of these methods in traffic safety, particularly in studies involving visual environment features extracted from street view imagery, remains at an early stage (Abécassis et al., 2025).

Table. 1 Representative traffic safety studies by visual environment features, causal inference, and effect heterogeneity.

| Study | Data source | Methodology | Safety outcome | Visual features | Causal Inference | Heterogeneity | Scale |
|---|---|---|---|---|---|---|---|
| (Wang et al., 2025) | Crash data | Probabilistic graphical models (GGM, CBN, XGBoost) | Grade crossing collision frequency | × | √ | × | At grade crossings in Canada |
| (Guo et al., 2024) | Crash data | Causal forest & DML | Traffic incident duration | × | √ | √ (CATE) | Urban transportation network of Tianjin |
| (Park and Lee, 2026) | NSV & Crash data | Deep learning (CNN) & RF, XGBoost, Gradient Boost | Traffic crash frequency | √ | × | × | Image-based street-level analysis in Seoul |
| (Ye et al., 2024) | GSV & Severity of cycle accident | Deep learning (CNN) & XGBoost | Cycling safety level of road environment | √ | × | △ (Road type) | Image-based analysis of the London road network |
| (Zhang et al., 2025) | Street view images & GPS-based vehicle trajectory data | Copula model | Driver distraction ratio & Speed standard deviation | √ | × | △ (Driver age) | Image-based driver behavior analysis in the Orlando City of Florida |
| (Naseralavi, 2025) | Rear-end collision crash records with injury severity | Binary logit model | Injury severity crash frequency | × | × | △ (Gender, age, area type) | Highway network of California |
| (Metz-Peeters, 2025) | Injury crash data | Causal forest | Injury crash frequency | × | √ | √ (CATE) | German motorway network |
| (Shen et al., 2025) | GSV & Crash data | Random forest (RF) | Frequency of elderly pedestrian crashes | √ | × | △(age) | Street-view image-based analysis of crash risk in Hong Kong |
| (Li et al., 2024) | Crash data | Doubly robust ML | Post-crash traffic speed | × | √ | √ (CATE) | On Interstate 5 in Washington State |
| (Asadi et al., 2022) | Urban traffic crash data | Negative Binomial regression model | Property damage only crash frequency | √ | × | △ (area type) | Area-level analysis in Dutch Randstad |
| (White and Meixler, 2024) | NAIP & Crash data | Difference-in-differences | Crash rates and fatal crash frequency | √ | √ | × | Statewide analysis along major interstate corridors in Georgia |
| (Cui et al., 2025) | GSV & Crash data | Multiscale geographical random forest | Crash frequency | √ | × | △ (Space) | Macro-level modeling across Southeast Florida |
| (Wang et al., 2026) | GSV, Crash data & connected vehicle data | ML prediction & Causal mediation analysis | Crash risk classification outcome | √ | √ | × | High-speed highway segments in Florida |

Note**:** √: explicitly modeled; △: partial / subgroup analysis; ×: not addressed; NSV: Naver Street View; GSV: Google Street View; NAIP: National Agriculture Imagery Program; CNN: Convolutional neural networks; GGM: Generalized Gamma Model, CBN: Conditional Bayesian Network, RF: Random Forest, ML: Machine Learning, DML = Double Machine Learning; CATE: conditional average treatment effects.

## 3 Data preparation

This study constructs a multisource spatial dataset to investigate traffic safety in southeastern Florida. By mapping crash data, socioeconomic indicators, and visual environment features onto a uniform grid, we address spatial inconsistency across data sources. The variable construction processes necessary to prepare these inputs for causal inference are then detailed.

### 3.1 Research area and crash frequency quantification

As illustrated in Fig. 2, the study area covers the region of Miami-Dade, Broward, and Palm Beach in southeastern Florida, which is characterized by persistently high traffic crash frequency and severity within the state. The analysis focuses on the urban and suburban corridors along the eastern coast, which concentrate the majority of the population, economic activity, and transportation demand, while sparsely populated western areas with minimal road infrastructure were excluded.

To mitigate data integration difficulties and potential biases regarding scale and boundaries caused by the irregular size, shape, and function of conventional spatial units such as census tracts, traffic analysis zones or road segments (X. Li et al., 2022). A uniform 2 km × 2 km grid system was constructed across the three counties, yielding 1,042 analytical units. This grid resolution balances the preservation of local heterogeneity in traffic and visual environments with the limitation of spatial spillover effects (Gan et al., 2025).

Traffic crash data were obtained from the Florida Department of Transportation's Signal Four Analytics database. The dataset includes geocoded records of reported traffic crashes from 2021 to 2023. Each crash event was spatially assigned to the predefined grid system based on its coordinates. For each grid i, $Y_I$ is calculated as the total number of crash events occurring within its spatial boundary, following standard practice in grid level traffic safety analysis (Mohammed et al., 2023; Sung et al., 2022):

$$Y_i = \sum_{k=1}^{K} \mathbb{I}(P_k \in G_I) \quad (1)$$

where $P_k$ is the k-th crash point, $G_i$ represents the geometric boundary of grid $i$, and $\mathbb{I}(P_k \in G_i)$ is an indicator function that equals 1 if crash point $P_k$ is located within grid $G_i$, and 0 otherwise.

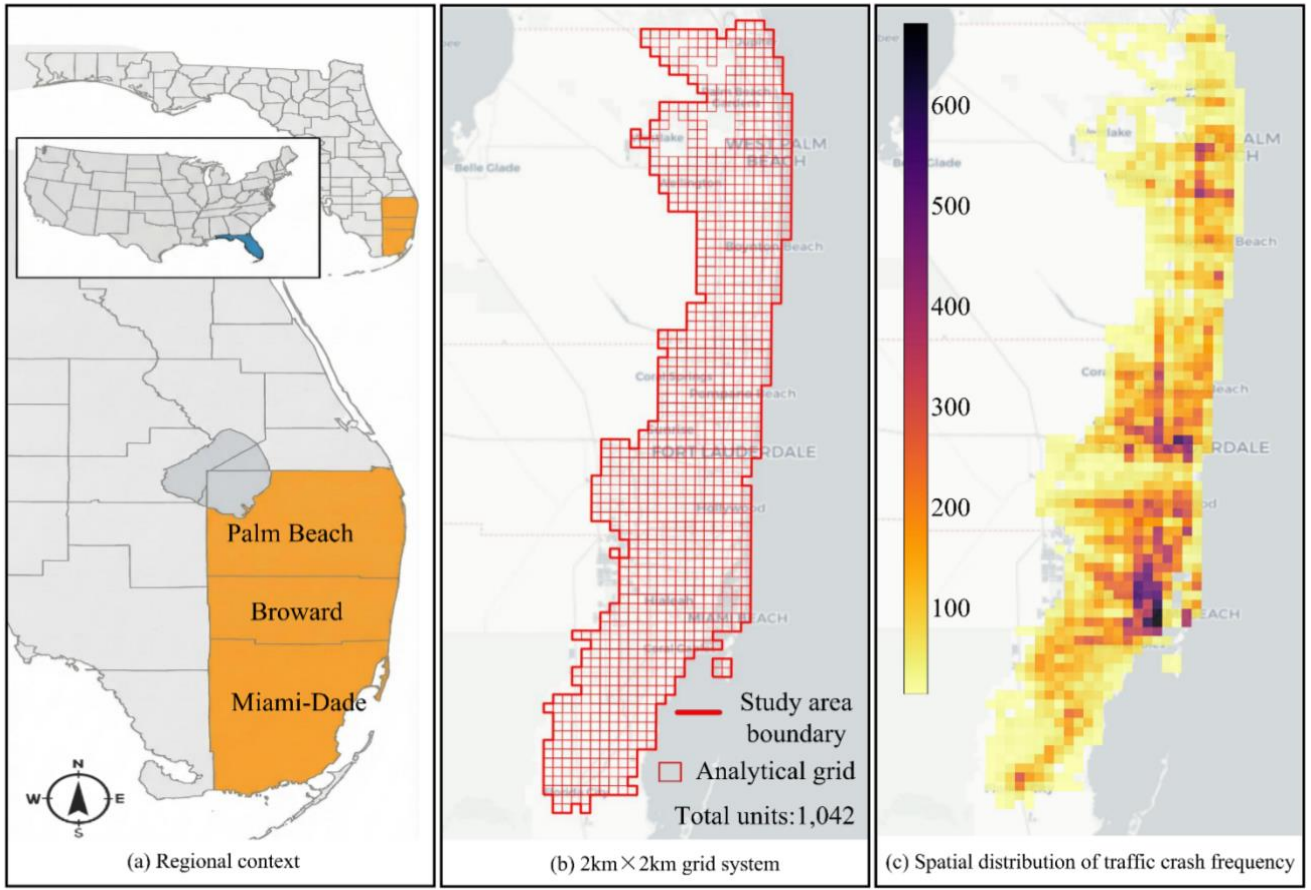

Fig. 2. Research area and spatial grid creation. (a) Location of Miami-Dade, Broward, and Palm Beach; (b) The uniform 2km×2km grid system covering the study area; (c) Traffic crash frequency by grid cell.

3.2. Data acquisition and variable construction

3.2.1 Acquisition and spatial integration of social vulnerability features

The socioeconomic variables were obtained from the U.S. social vulnerability index (SVI). As this dataset is provided at the census tract level, a spatial transformation was necessary to match our grid system. As shown in Fig. 3 (a), a weighting scheme based on area and population was used to redistribute social vulnerability index features to each grid cell, thereby achieving this correspondence. (Cui et al., 2025).

3.2.2 Extraction and quantification of visual streetscape features

Visual environment features extracted from Google Street View imagery (GSV). As illustrated in Fig. 3 (b), we generated 57,088 sampling points at 100 m intervals along the road network. At each point, four images (0°, 90°, 180°, 270°) were captured with a 50° horizontal field of view and 0° pitch to form 360° panoramas, yielding 228,352 images in total (Li et al., 2015). Semantic segmentation was performed using the Segmenter Transformer model (Strudel et al., 2021) to classify pixels into 18 categories (e.g., road, building, greenery). The proportional coverage of each category was calculated as the feature input. Additionally, visual complexity was quantified via visual entropy (H) (Guan et al., 2022).

$$H_i = -\sum_{k=1}^{18} p_k \log_2(p_k) \tag{2}$$

where $p_k$ is the pixel proportion of the k-th object category in the image, and $H_i$ is the resulting entropy value, effectively capturing the complexity and informational load of the visual environment. These image level metrics were then aggregated to the grid level by averaging, resulting in a final set of 19 visual environmental features.

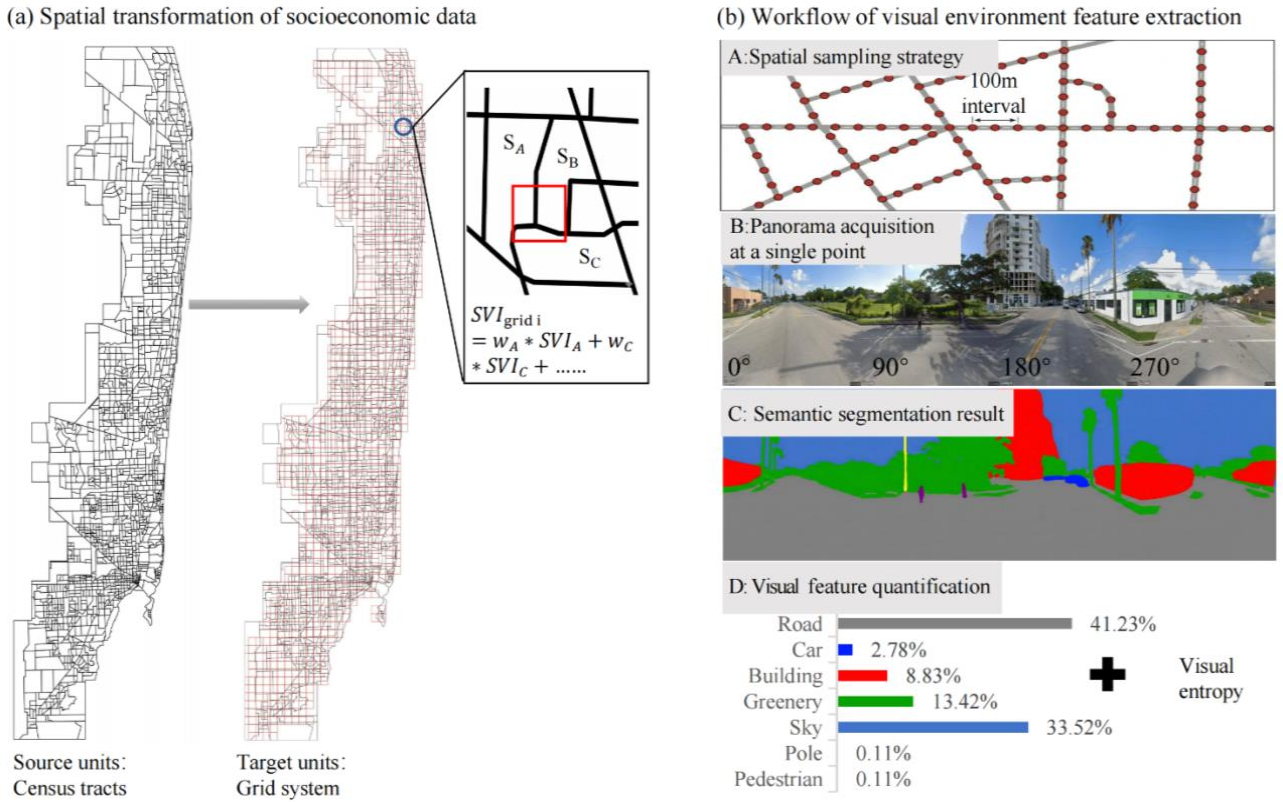

Fig. 3 Spatial transformation of socioeconomic data and visual environment feature extraction.

3.3 Variable setting for causal analysis

3.3.1 Conceptual classification of variables

Guided by the potential outcome's framework, study variables were explicitly defined to support causal identification of the effects of micro-scale visual environment features on traffic crash. Consistent with standard practice in causal inference, all variables were organized into three conceptually distinct categories outcome, treatment, and confounders to facilitate transparent estimation and interpretation of causal effects (Feuerriegel et al., 2024).

The traffic crash frequency within each grid cell serves as the outcome variable. Visual environment features derived from street view imagery are considered candidate explanatory variables, with each feature examined individually as a treatment in the causal analysis. As shown in the Fig. 4, all remaining visual environment features and socioeconomic indicators are treated as covariates to account for confounding influences. This formulation enables the estimation of the causal effect of a specific visual attribute while conditioning on a rich set of contextual characteristics.

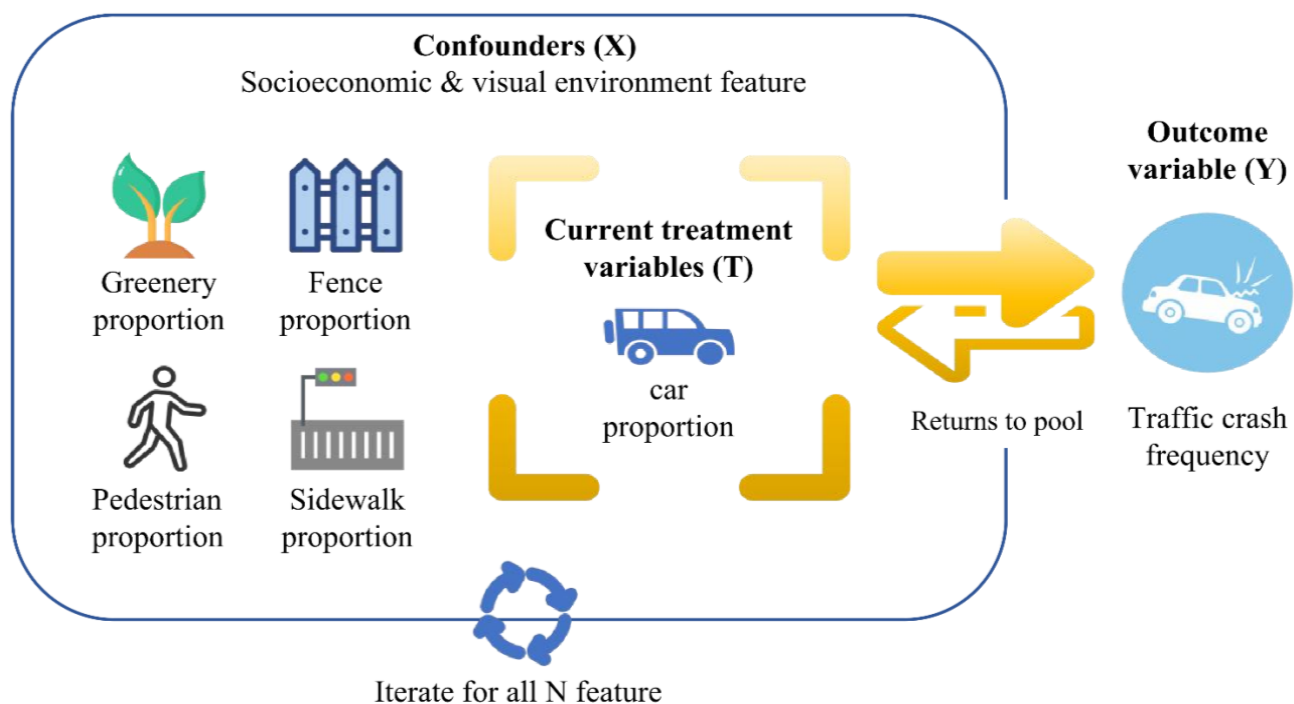

Fig. 4 Causal effect estimation workflow with rotating treatment features and shared confounders.

3.3.2 Covariate screening and dimensionality reduction

Satisfying the conditional independence assumption requires a sufficiently rich covariate set, yet high-dimensional features may introduce noise and inflate model variance (Cheng et al., 2022). To balance these considerations, a screening procedure was applied to identify a parsimonious set of relevant covariates. Spearman rank correlations between candidate covariates and crash frequency were first examined, and the 17 variables with the largest absolute correlations were retained (Table. 2). While correlation alone does not establish causality, it provides a preliminary indication of features likely to lie on the causal pathway or act as confounders. The retained variables were then evaluated using an XGBoost model to account for potential nonlinear relationships and interactions.

Table. 2 Descriptive statistics of the selected candidate covariates.

| Feature | Description | Min | Max | Mean | Std |
|---|---|---|---|---|---|
| Socioeconomic features | | | | | |
| Population | Population size in study grid | 3.99 | 35219.42 | 5909.83 | 4649.81 |
| No vehicle | Percentage of households with no vehicle available (%) | 0.00 | 42.04 | 5.53 | 5.01 |
| Poverty below 150 | Percentage of person below 150 % poverty (%) | 1.50 | 70.56 | 18.88 | 10.63 |
| Crowding | Percentage of occupied housing units with more people than rooms (%) | 0.00 | 23.57 | 4.33 | 3.61 |
| Limited English | Percentage of person (age 5+) who speak English "less than well" (%) | 0.00 | 51.32 | 9.89 | 9.23 |
| Housing burden | Percent of households with high housing cost burden (%) | 5.99 | 63.15 | 30.94 | 11.77 |
| Uninsured | Percent of persons without health insurance (%) | 0.00 | 41.44 | 12.30 | 7.03 |
| Minority | Percent of persons who are racial/ethnic minority (%) | 2.80 | 99.68 | 61.44 | 26.55 |
| No high school | Percent of persons (age 25+) with no high school diploma (%) | 0.00 | 52.33 | 11.00 | 8.51 |
| Visual environment features | | | | | |
| Car proportion | The mean car proportion in GSV images | 7.45e-06 | 0.13 | 0.01 | 0.01 |
| Greenery proportion | The mean terrain (grass/ greenery) proportion in GSV images | 0.01 | 0.25 | 0.11 | 0.04 |
| Fence proportion | The mean fence proportion in GSV images | 0.00 | 0.07 | 0.01 | 0.01 |
| Pedestrian proportion | The mean person proportion in GSV images | 0.00 | 0.01 | 2.801e-04 | 4.970e-04 |
| Sidewalk proportion | The mean sidewalk proportion in GSV images | 1.65e-05 | 0.08 | 0.02 | 0.01 |
| Road proportion | The mean road proportion in GSV images | 0.16 | 0.43 | 0.30 | 0.04 |
| Pole proportion | The mean pole proportion in GSV images | 3.61e-05 | 0.01 | 4.81e-03 | 1.99e-03 |
| Building proportion | The mean building proportion in GSV images | 0.00 | 0.30 | 0.02 | 0.03 |

By jointly considering correlation strength and feature importance, a final set of 10 covariates comprising five socioeconomic indicators and five visual environment features was selected for causal analysis. All continuous variables were standardized using Z-score normalization prior to estimation. The spatial distributions of the selected variables are illustrated in Fig. 5.

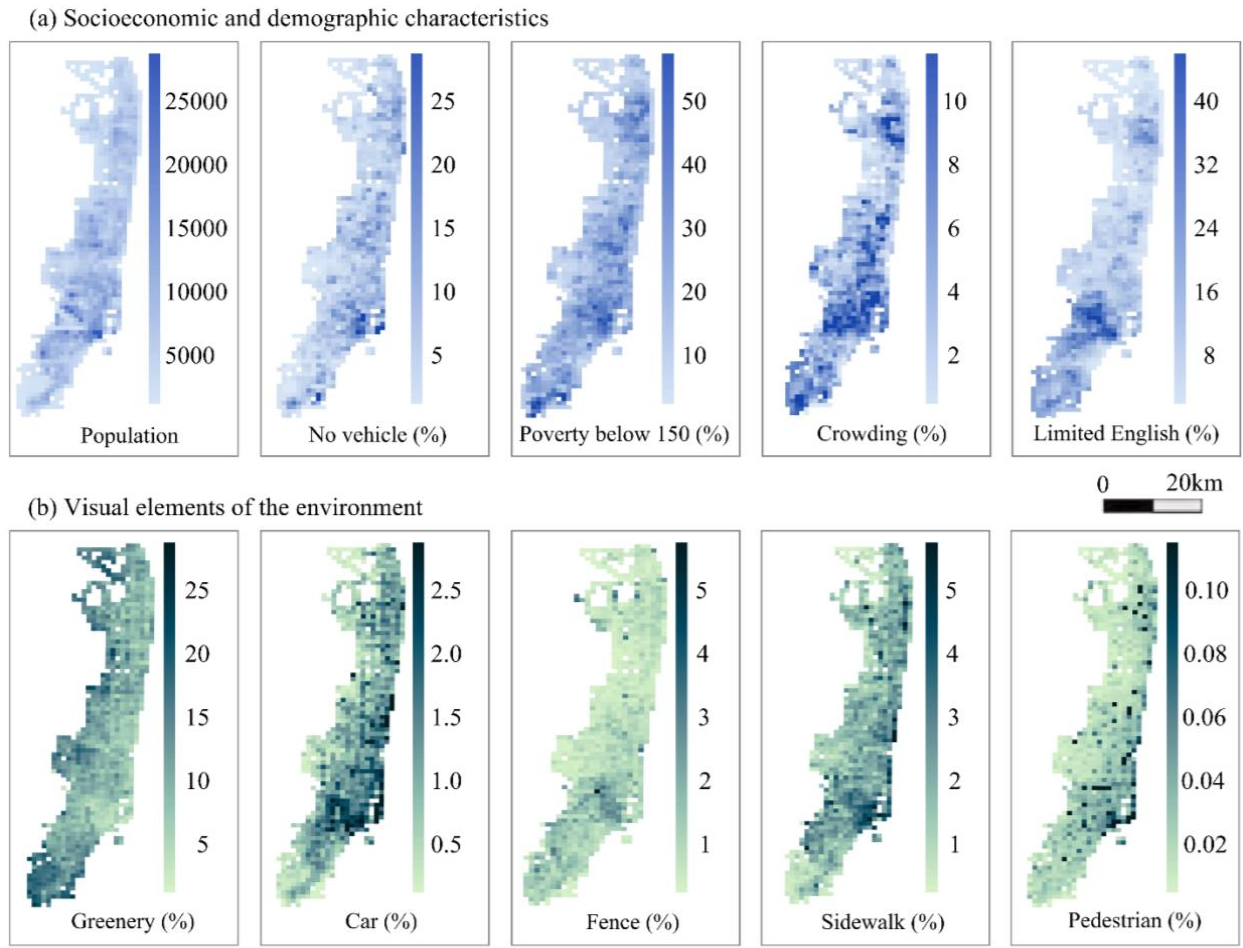

Fig. 5 Geospatial patterns of the identified key environmental and socioeconomic confounders.

## 4. Methodology

Given the lack of conclusive evidence regarding the causal relationships between specific visual environment features and traffic crashes, the proposed methodology adopts a causal inference design comprising two-stage, as illustrated in Fig. 6. The first stage focuses on the robust identification of key features that exhibit causal effects on traffic crashes, while the second stage examines heterogeneity in the effects of the selected feature.

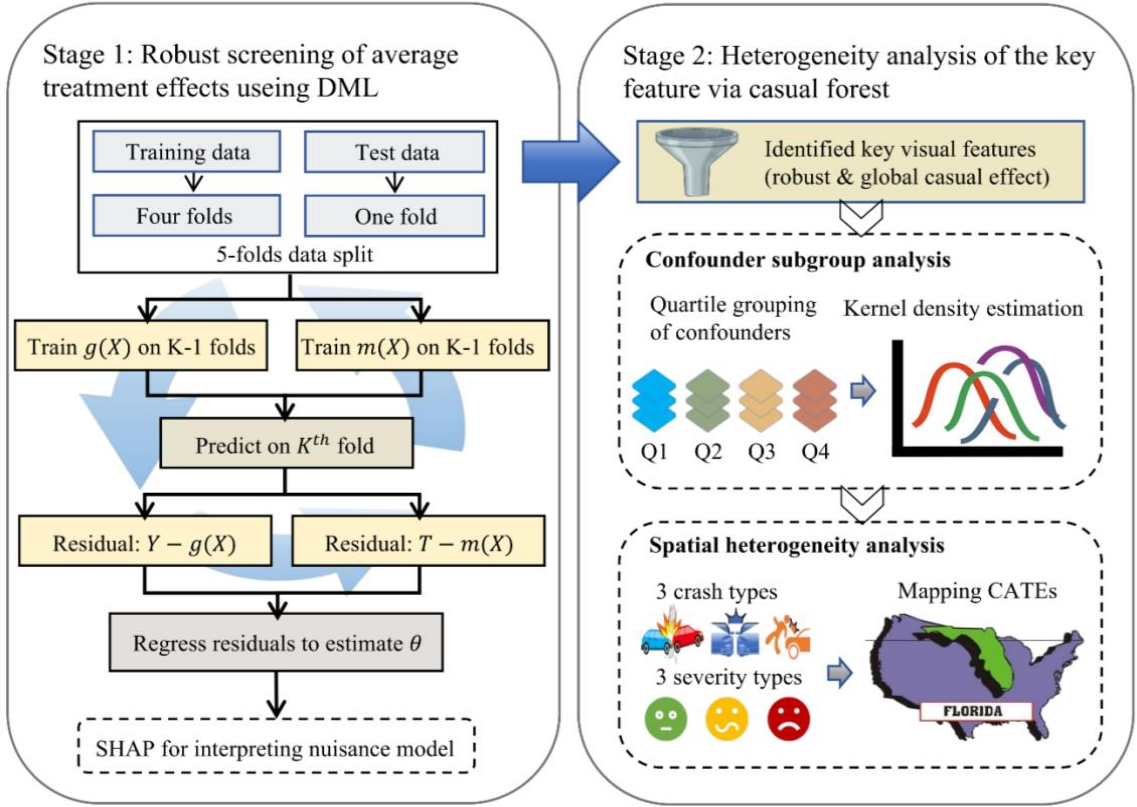

Fig. 6. Two-stage causal inference framework for identifying robust street view semantic effects and characterizing heterogeneity in traffic safety.

### 4.1. Causal identification assumptions

The implementation of this stepwise approach relies on a set of causal identification assumptions, adapted for our spatial context and continuous treatments (Hirano and Imbens, 2004). First, the conditional independence assumption requires that treatment assignment be independent of potential outcomes given the observed covariates. This study strengthens the plausibility of this assumption by incorporating a comprehensive set of socioeconomic and visual context measures, which jointly account for major sources of confounding between the visual environment and traffic safety.

Second, the overlap assumption necessitates sufficient common support across the treatment distribution (Imbens and Wooldridge, 2009). In practical terms, this implies that for any observable combination of contextual characteristics, there exists a meaningful range of variation in the treatment variable. Our estimation strategy explicitly respects this requirement by focusing inference on the region of common support, thereby avoiding excessive reliance on extrapolation derived from models and ensuring that causal comparisons are grounded in local empirical evidence.

Third, we maintain the stable unit treatment value assumption (SUTVA) which posits that the outcome of any given spatial unit depends solely on its own treatment assignment and remains unaffected by treatments assigned to other units (Fortuin, 2022). While spatial dependencies pose a potential threat to this assumption, our research design mitigates such concerns through the use of an optimized spatial grid scale that balances local relevance with minimal interference (Chen et al., 2020), coupled with high-dimensional controls that capture the structural factors underlying spatial autocorrelation processes.

4.2. Stage one: Robust screening of causal signals and interpretability analysis

4.2.1. Construction of the double machine learning estimator for global parameters

Given that the initial analysis targets global ATEs, we adopt a partially linear model that represents the causal effect as a single estimable coefficient (Chernozhukov et al., 2018). This formulation isolates the average effect of each visual feature while allowing other factors to influence crash outcomes in a flexible, potentially nonlinear manner (Bach et al., 2024). The model is defined by two structural equations:

$$Y = \theta T + g(X) + \varepsilon, E[\varepsilon|X,T] = 0 \tag{3}$$

$$T = m(X) + v, E[v|X] = 0 \tag{4}$$

It isolates the causal parameter $\theta$ for the treatment effect from the high-dimensional confounding controls, which are handled by flexibly estimating the nuisance functions $g(X)$ and $m(X)$ using machine learning.

4.2.2. Estimation procedure and model interpretation

We estimate $\theta$ using a 5-fold cross fitting algorithm. First, we obtain residuals for the outcome and treatment by subtracting the predicted values from $Y$ and $T$, respectively, and then regress the resulting outcome residuals on the treatment residuals. Implementation of this partially linear model utilized four high performance learners, including Random Forest (Cutler et al., 2012), XGBoost (Chen and Guestrin, 2016), Gradient Boosting Machine (GBM) (Friedman, 2001), and LightGBM (Ke et al., 2017), ensuring our ATE estimates are not an artifact of a single algorithm.

To interpret the nuisance functions fitted by the machine learning models within the DML framework, we employed the SHAP for model interpretation (Lundberg and Lee, 2017). SHAP quantifies the marginal contribution of each covariate to the model's prediction, providing a clear understanding of how the nuisance functions capture the effects of different features. For each feature $j$, its SHAP value $\phi_j$ reflects the positive or negative influence of that feature on the prediction for a given instance. Formally, for a predictive model $f$ and instance $x$, the SHAP value is defined as:

$$\phi_j(f,x) = \sum_{S \subseteq \{1,\dots,p\}\setminus\{j\}} \frac{|S|!(p-|S|-1)!}{p!} \left(f_x(S \cup \{j\}) - f_x(S)\right) \qquad (5)$$

where $S$ is a subset of features excluding $j$, $p$ is the total number of features, $|S|$ is the size of subset $S$, and $f_x(S)$ denotes the model prediction for instance $x$ using only the feature subset $S$. This formulation ensures a fair allocation of the difference between the actual prediction and a baseline prediction among all features.

Global feature importance rankings were obtained by computing the mean absolute SHAP value of each feature across all samples, providing a relative measure of each covariate's contribution to the model predictions (Li et al., 2023; Yang et al., 2025). SHAP dependence plots were generated to examine the relationship between feature values and their contributions, with a LOWESS curve superimposed to illustrate nonlinear patterns in the model output.

4.3. Stage two: Nonparametric modeling and multidimensional heterogeneity analysis

4.3.1. Causal Forest specification for continuous treatment heterogeneity

The second stage focuses on heterogeneous treatment effects for the key visual feature identified in the first stage, which exhibited a globally robust and consistent causal impact. While the first stage provided an overall assessment of feature effects, it does not capture potential variation across different environments or crash types. To address this,, we employ a Causal Forest model, a generalized random forest method adapted for continuous treatments, to nonparametrically estimate the CATE (Wager and Athey, 2018). Within the potential outcomes framework for continuous treatments, the target causal parameter is defined as the conditional average marginal effect (Lee et al., 2025; Souto and Neto, 2024), representing the instantaneous causal effect of a marginal increase in the treatment for a unit with covariates $X = x$:

$$\tau(x) = \frac{\partial \mathbb{E}[Y \mid T = t, X = x]}{\partial t} \qquad (6)$$

The Causal Forest estimates $\tau(x)$ by growing an ensemble of honest regression trees. Honesty is enforced by using separate subsamples for determining the splits of each tree and for estimating the effects within its leaves, which reduces overfitting and ensures asymptotic unbiasedness (Van Vogt et al., 2025). The forest recursively partitions the covariate space with the objective of maximizing heterogeneity in the estimated treatment effects $\tau(x)$ across different nodes. Computationally, within each tree neighborhood, the estimation is based on a locally linear approximation. For a target covariate profile $x$, the Causal Forest solves a weighted least squares problem centered around the local mean of the treatment, yielding the estimator:

$$\hat{\tau}(x) = \frac{\sum_{i=1}^{n} w_i(x)(T_i - \bar{T}_w(x))(Y_i - \bar{Y}_w(x))}{\sum_{i=1}^{n} w_i(x)(T_i - \bar{T}_w(x))^2} \quad (7)$$

where $w_i(x)$ is the weight assigned to observation $i$ based on its proximity to $x$ in the covariate space, and $\bar{T}_w(x)$, $\bar{Y}_w(x)$ are the corresponding locally weighted means. This formulation shows that the Causal Forest performs a nonparametric regression of the residualized outcome on the residualized treatment, using a adaptive kernel determined by the forest structure (Athey et al., 2019).

4.3.2. Spatial mapping and stratified analysis of conditional treatment effects

The resulting CATE estimates were linked to their geographical coordinates, producing distribution maps that depict the spatial variation in the impact of visual environment features across urban areas (Credit and Lehnert, 2024). To further investigate treatment effect heterogeneity, both heterogeneity dependent on covariates and heterogeneity specific to outcomes were considered. Continuous confounders were discretized into quartile subgroups to estimate subgroup ATEs, and kernel density curves of CATEs were plotted for each group. This allowed for an examination of treatment effect variations across different covariate profiles, revealing heterogeneity that may be masked in aggregate estimates (Komura et al., 2025). Finally, for each of six crash subtypes representing distinct severity levels and crash types, a dedicated Causal Forest model was estimated to assess spatial variation in the visual feature's effect.

5. Results and discussion

This chapter presents the findings of our two-stage causal inference framework, structured to progress from global causal identification to local heterogeneity analysis. First, Section 5.1 employs Double Machine Learning (DML) to conduct robust screening, identifying the proportion of greenery as the sole visual environment feature with a statistically significant negative impact on crash frequency. Building on this global baseline, Section 5.2 utilizes SHAP analysis to uncover the nonlinear mechanisms driving these safety effects. Subsequently, the analysis deepens in Section 5.3, where the Causal Forest model quantifies multidimensional heterogeneity across socioeconomic, crash-type, and geospatial dimensions. Finally, Section 5.4 corroborates these quantitative results through a comparative analysis of representative street view cases.

5.1 Global causal identification and estimator validation

5.1.1 Robust selection of the primary treatment variable

While theoretical frameworks posit various visual environment effects on traffic safety, translating these constructs into empirical evidence requires rigorous validation. Specifically, it is critical to identify which visual elements possess statistically robust causal validity before conducting in-depth heterogeneity analysis. To ensure the reliability of the treatment variable, we must confirm the cross-model robustness of the Average Treatment Effect (ATE). This step is essential to verify global causal relationships while mitigating biases specific to any single algorithm. Accordingly, this study performed multi-model robustness tests on five standardized visual environmental variables (as illustrated in Fig. 7). The results indicate that among the features examined, only the greenery proportion exhibited consistent and significant negative causal effects across all four machine learning algorithms. The estimated ATEs ranged from -5.12 to -7.43 ($p < 0.05$). This finding aligns with prior evidence suggesting that street level greenery is associated with improved traffic safety outcomes, likely through visual traffic calming and behavioral adaptation mechanisms (Zhu et al., 2022). Given its strong and robust causal signal, the greenery proportion was selected as the primary treatment variable for the subsequent heterogeneity analysis.

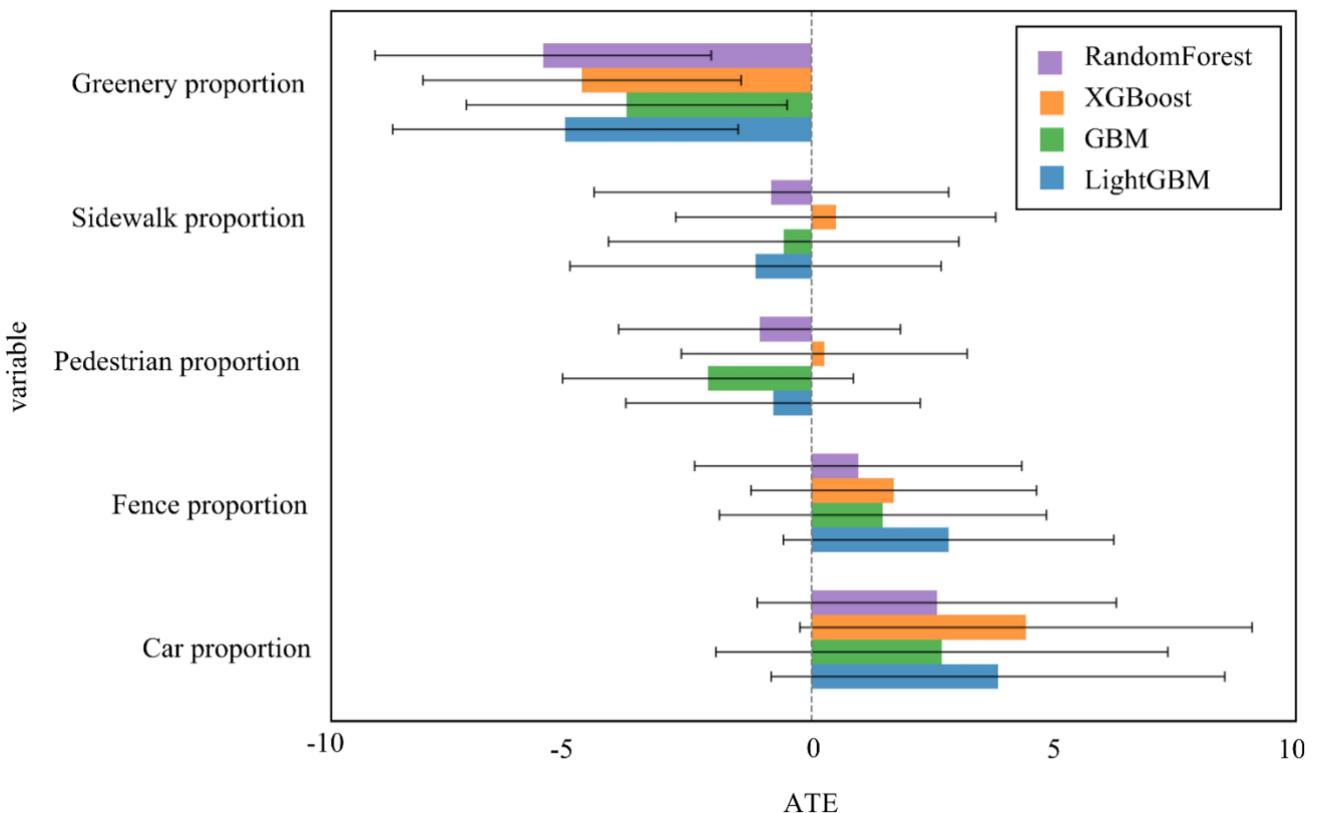

Fig. 7. Robustness check for ATEs and 95% confidence intervals of different visual environment features.

5.1.2 Model validation and comparative benchmarking

To ensure the rigorousness of the estimated treatment effects, we conducted a comprehensive validation of the DML estimator, evaluating both its internal stability and external consistency. Table. 3 summarize the performance metrics and comparative results, providing a holistic view of the estimator's reliability. First, we assessed the robustness of the DML estimator against the stochastic variability of sample splitting. Following Bayesian hyperparameter optimization (Snoek et al., 2012), we performed ten independent replications of five-fold cross-validation. As detailed in Table. 3, the estimated ATE of the greenery proportion remained remarkably consistent, with a mean value of -6.38 (95% CI: -10.69 to -2.08, p = 0.005). Furthermore, we benchmark the DML estimates against Ordinary Least Squares (OLS) and Causal Forest models to contextualize the magnitude of the effect. While all three methods consistently identify a significant negative impact of greenery on crash frequency, the comparison reveals critical methodological distinctions. The OLS model produced the largest effect size (-13.52). This substantial deviation likely reflects upward bias caused by unobserved nonlinear confounding that linear models fail to capture (Asadi Ghalehni and Afghari, 2026). The Causal Forest yielded the ATE of -8.94, while closer to the DML result, this estimate carries a larger standard error of 4.54 compared to 2.20 for the DML model. This reduction in precision is expected, as Causal Forests are designed to estimate treatment effect heterogeneity at the unit level rather than a single global parameter, reflecting the greater flexibility and complexity inherent to nonparametric estimation. Given this consistency across methods, we retain the DML estimate as the primary and more precise benchmark for the average effect and employ the Causal Forest to examine granular treatment effect heterogeneity in the subsequent section.

Table. 3 Comparative assessment of average treatment effects across different estimation strategies.

| Model | ATE | SE | 95% CI | P value |
| --- | --- | --- | --- | --- |
| DML-PLR | -6.38 | 2.20 | [-10.69,-2.08] | 0.005 |
| Causal Forest | -8.94 | 4.54 | [-17.84,-0.05] | 0.048 |
| Baseline (OLS) | -13.52 | 2.97 | [-19.34, -7.69] | 0.001 |

Note: SE = Standard Error; CI = Confidence Interval.

5.2 Mechanistic analysis of confounding features

5.2.1 Dominant determinants and feature interaction analysis

To validate the model's logical consistency and uncover latent confounding mechanisms, we utilized SHAP analysis to quantify both the primary features and their interactions as shown in Fig. 8. This unified assessment reveals distinct mechanistic pathways for the treatment and outcome models. As illustrated in Fig. 8 (a) and Fig. 8 (c), the analyses reveal the dominant determinants for the treatment and outcome models, respectively. For greenery distribution, the car proportion and sidewalk proportion emerge as the primary predictors, highlighting that the rigid allocation of street space and physical infrastructure constraints serve as the fundamental determinants of visible greenery (Liu et al., 2024). Conversely, traffic crash frequency is shown to be decisively driven by population size and car proportion, confirming that the outcome model successfully prioritizes exposure intensity and the scale of conflict opportunities rather than static environmental design alone (Høye and Hesjevoll, 2020). Furthermore, Fig. 8 (b) and Fig. 8 (d) visualize the interaction networks of confounding features, bridging this gap in perspective by mapping the absolute magnitude of SHAP main effects to node sizes and pairwise interaction strengths to edge thicknesses. For example, the most significant interaction effects on traffic crash occurs between the car proportion and the sidewalk proportion as shown in Fig. 8 (d). Given the finite nature of street space, this strong interaction serves as a quantitative proxy for the competitive tradeoff between motor vehicle dominance and pedestrian infrastructure, a pattern consistent with urban design research (Gerike et al., 2021).

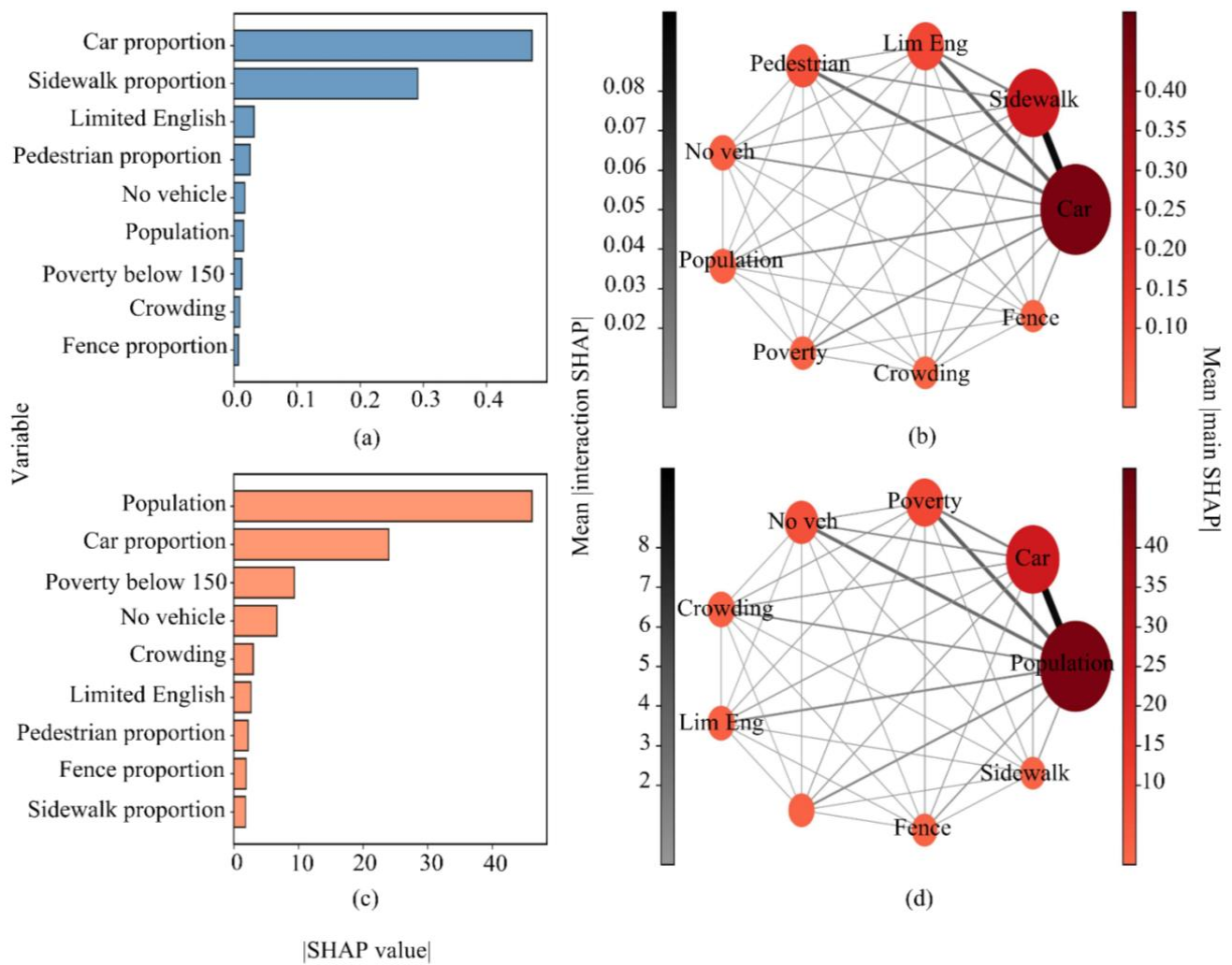

Fig. 8. SHAP-based identification of dominant determinants and interaction pathways. (a) Key predictors of greenery distribution; (b) Interaction network effects on greenery distribution; (c) Key predictors of crash frequency; (d) Interactions network effects on crash frequency.

5.2.3 Threshold effects and nonlinear dependencies analysis

While feature importance rankings isolate the primary determinants, they do not fully capture the non-monotonic nature of these relationships. To address this, SHAP dependence plots were employed to characterize the nonlinear mechanisms and threshold effects of key confounders. As illustrated in Fig. 9, both car and sidewalk proportions display distinct nonlinear patterns, transitioning from positive contributors to restrictive constraints once their intensity surpasses a critical threshold. This trend reflects the physical competition within the finite street envelope, where extensive allocation to transport infrastructure—whether for vehicular or pedestrian use—inherently limits the space available for ecological features. A similar constraint is evident in socioeconomic variables; high poverty rates and transit reliance in compact urban cores tend to restrict potential green infrastructure, driven by the prioritization of essential mobility needs over environmental amenities.

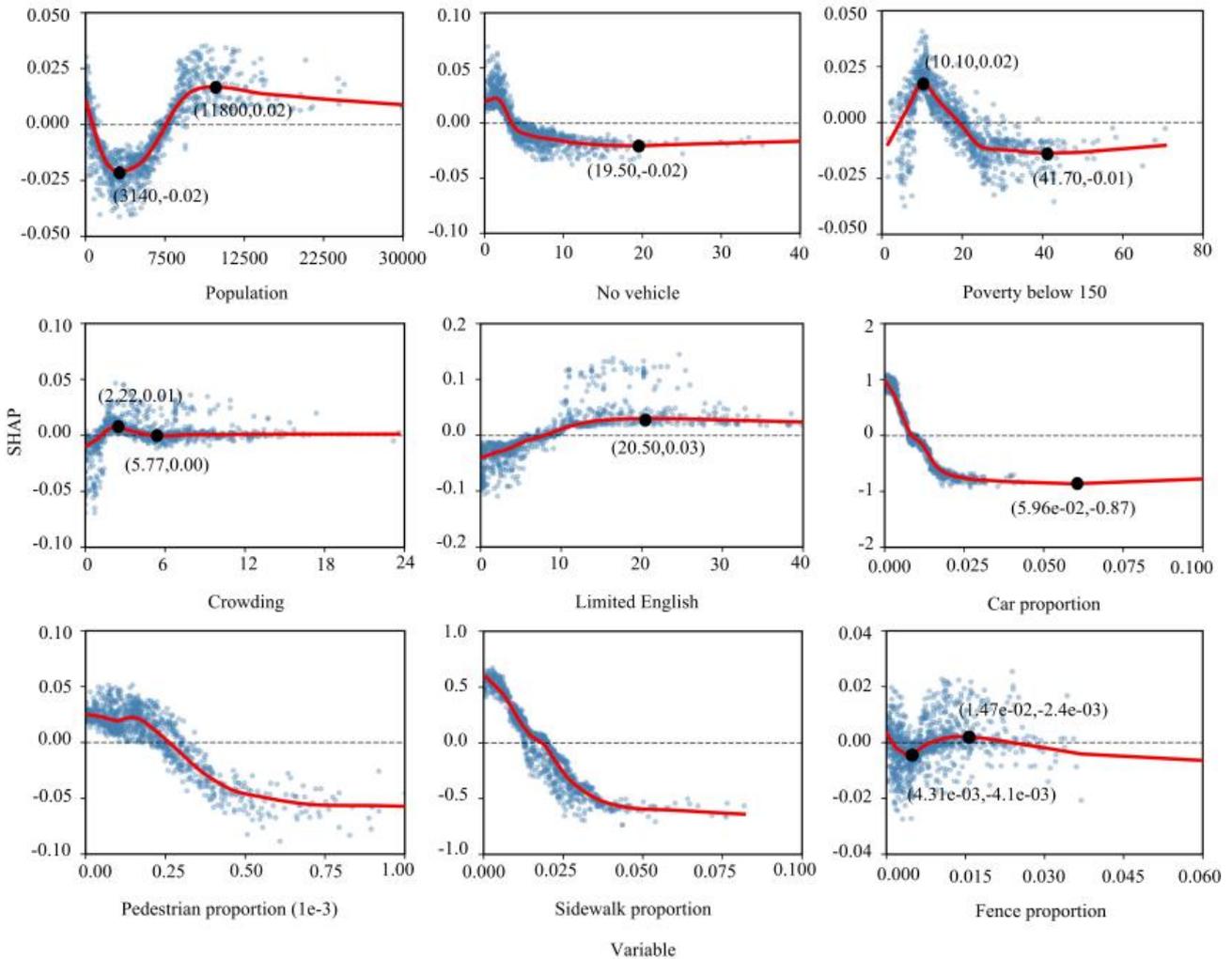

Fig. 9. Nonlinear effects on greenery allocation revealed by SHAP dependence plots.

Fig. 10 reveals the nonlinear effects of confounding variables on traffic crash frequency.

According to the population size and car proportion curves, traffic crash frequency is most sensitive when population size and car density rise rapidly. This highlights the importance of early intervention for traffic and street design management. High proportions of no vehicle households serve as a proxy for vulnerable road users, who are more frequently exposed to traffic hazards through walking or public transit. This underscores that traffic safety interventions in these communities should extend beyond engineering measures, integrating infrastructure improvements and public services.

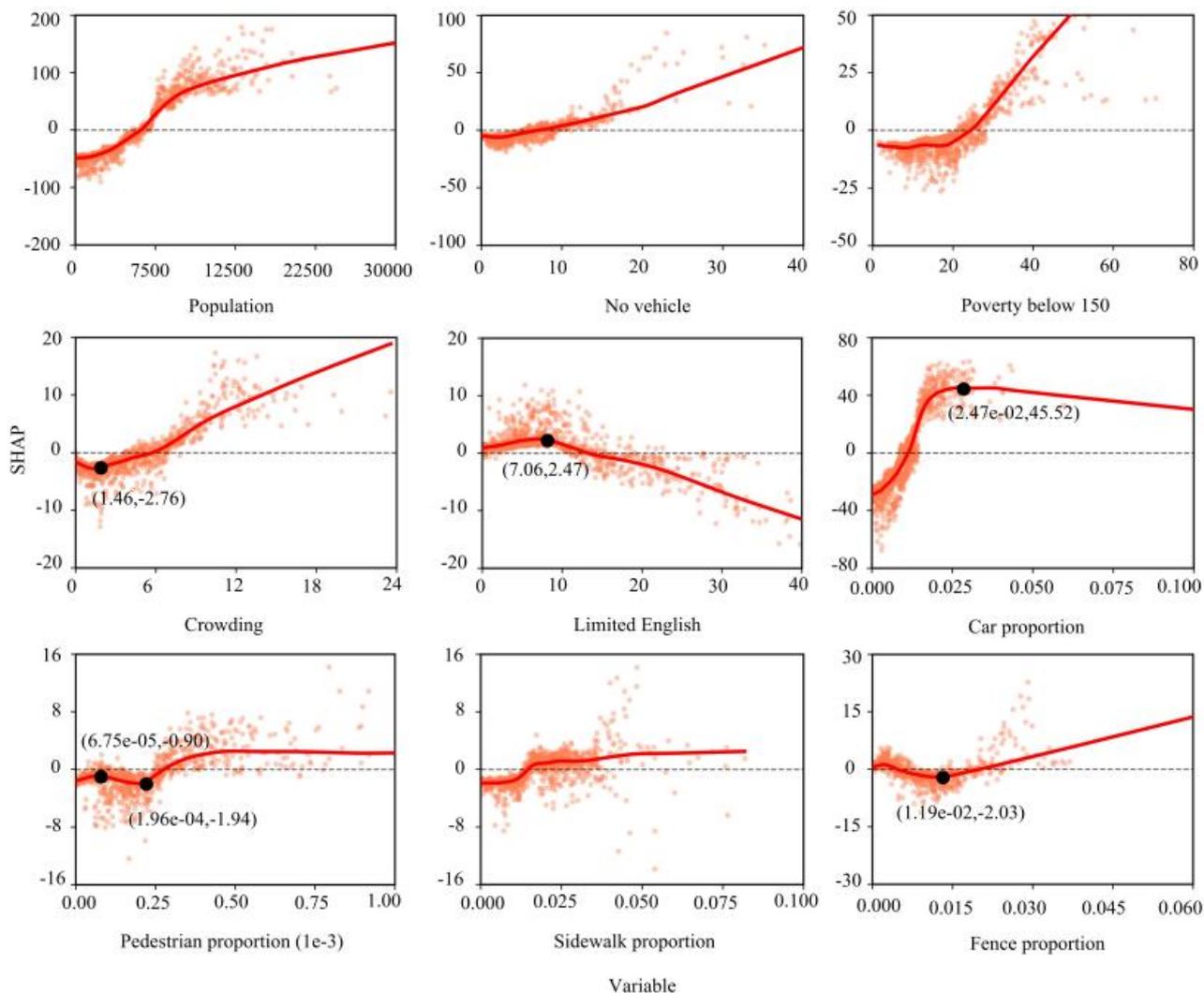

Fig. 10. Nonlinear effects on crash frequency revealed by SHAP dependence plots.

5.3 Multidimensional heterogeneity analysis using causal forests

5.3.1 Group analysis of effect magnitude and stability

While the ATE confirms the global protective role of street-level greenery, this aggregate measure masks significant spatial heterogeneity within complex urban systems. By leveraging the Causal Forest model to estimate Conditional Average Treatment Effects (CATE), we stratified key confounding variables into quartiles (Q1–Q4) to dissect how safety benefits vary across distinct physical and socioeconomic contexts. The resulting probability density distributions are presented in Fig. 11. A clear difference in effect magnitude is observed across strata. For socioeconomic dimensions, the average protective effect in the highest quartile (Q4) is approximately 3–4 times stronger than in the lowest quartile (Q1). This indicates that the protective role of greenery is significantly more pronounced in communities characterized by high poverty and residential density. A similar pattern applies to visual environmental features: greenery provides greater safety benefits in urban areas with high pedestrian and vehicular flows, whereas its effect is comparatively limited in low-density suburban environments. This stable grouping pattern likely reflects latent moderators, such as urban development intensity or socioeconomic disadvantage, pointing to directions for future research (Zaidi et al., 2025).

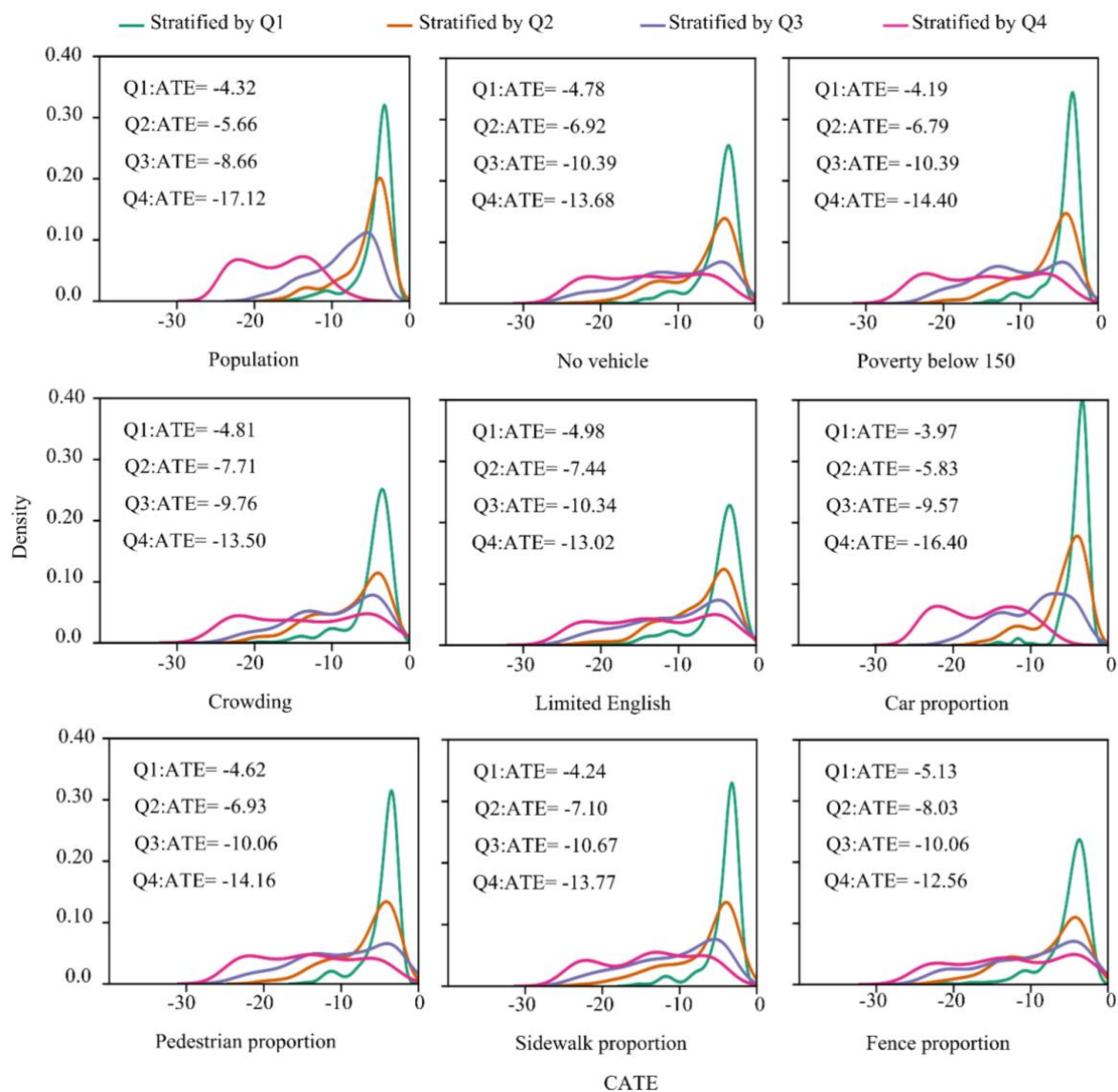

Fig. 11. Shifting patterns of effect magnitude and uncertainty across diverse urban contexts.

### 5.3.2 Comparative assessment of marginal safety benefits

Building on the finding that greenery efficacy peaks in resource-scarce urban cores (Section 5.3.1), this section quantifies the marginal safety benefits across specific crash configurations to support targeted policymaking. Given that establishing new greenery yields higher marginal returns in coverage-scarce areas, we employ a semi-elasticity metric (Δ%) to standardize the comparison. This metric anchors the analysis to neighborhoods with maximum scarcity (Q1), estimating the proportional improvement in safety outcomes per unit standard deviation increase in greenery. Table. 4 presents the estimated Average Treatment Effects (ATE) and marginal benefits. The results indicate that fatal, rear-end, and serious injury crashes exhibit the highest marginal efficiencies, with semi-elasticity values of 11%, 9%, and 8%, respectively. This suggests that green infrastructure is particularly efficient at mitigating high-severity outcomes. However, policy interpretation must account for baseline crash levels denoted by $\bar{Y}(Q1)$. Angle and injury crashes exhibit considerably higher baselines in scarce greenery areas (61.55 and 142.14), implying that protective effects for these types should not be overlooked. Overall, Δ% should be interpreted as a measure of proportional benefit rather than a direct indicator of absolute policy effectiveness, and it should be evaluated in conjunction with local crash incidence levels.

Table. 4 Average causal effects and marginal safety benefits of greenery across crash severities and types.

| Feature | Description | ATE | 95% CI | P value | $\bar{Y}$ | $\bar{Y}(Q1)$ | Δ% |
|---|---|---|---|---|---|---|---|
| Category Angle | Number of angle crashes | -3.29 | [-6.15, -0.43] | 0.024 | 34.79 | 61.55 | -5% |
| Category Pedestrian/Bicycle | Number of pedestrian & bicycle crashes | -0.76 | [-1.49,-0.02] | 0.045 | 7.65 | 16.00 | -5% |
| Category Rear End | Number of rear end crashes | -3.52 | [-5.98,-1.06] | 0.005 | 23.50 | 39.29 | -9% |
| Severity Fatality | Number of fatal crashes | -0.30 | [-0.56,-0.05] | 0.020 | 1.52 | 2.78 | -11% |
| Severity Serious Injury | Number of serious injury crashes | -0.74 | [-1.29,-0.19] | 0.008 | 5.32 | 9.62 | -8% |
| Severity Injury | Number of injury crashes | -7.96 | [-15.76,-0.16] | 0.045 | 79.74 | 142.14 | -6% |

N=1042
Terrain mean <= 0.0751 (Q1)

Notes: Δ% represents the semi elasticity indicating the percentage change in the outcome $\bar{Y}(Q1)$ for a one unit standard deviation increase in greenery. $\bar{Y}$ is the sample average of the outcome variable; $\bar{Y}(Q1)$ is the average traffic crash frequency value for the subgroup where the treatment variable greenery is below the 25th percentile.

### 5.3.3 Spatial heterogeneity of safety benefits across diverse crash typologies

To translate causal insights into actionable planning interventions, it is essential to precisely pinpoint the spatial loci of these effects. By jointly analyzing the CATE values across different crash typologies and their spatial distributions (Fig. 12), this section identifies the specific geographic units where the protective effects of greenery are most pronounced.

A consistent spatial pattern emerges across all crash types shows that the strongest protective effects are heavily concentrated within the three primary urban cores. This spatial clustering suggests that green infrastructure functions most efficiently in dense, high-activity zones. From a policy perspective, this confirms that directing investments toward urban cores—rather than spreading resources uniformly—represents the optimal strategy for maximizing aggregate crash reduction. Table. 4 presents the ATE estimates produced by the causal forest for the greenery proportion and traffic crash indicators. Analyzed by severity, greenery appears most effective in mitigating injury crashes (ATE = -7.96), likely due to the high baseline frequency of such incidents in complex urban environment. Regarding crash type, the protective effects are most pronounced for rear-end and angle crashes. This aligns with the mechanism that greenery reduces visual monotony and calms traffic speeds, thereby addressing the primary causes of vehicle-to-vehicle conflicts. The ATE for pedestrian and bicycle crashes is only −0.76. This disparity suggests that urban greenery in its current form—while effective for vehicular safety—provides limited protection for vulnerable road users. This finding highlights a critical gap in current design practices, indicating that simply increasing green coverage is insufficient for non-motorized safety without integrating specific protective infrastructure.

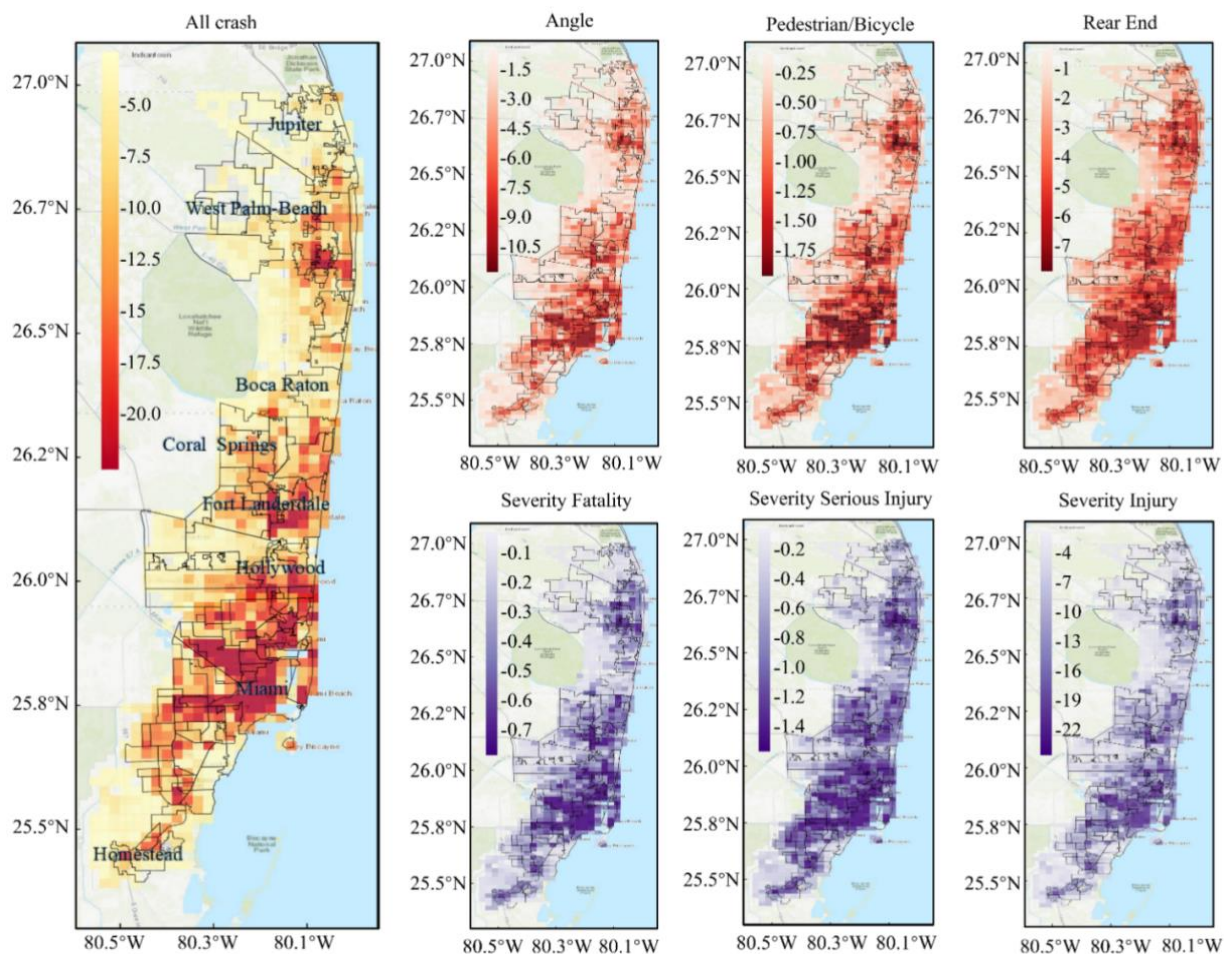

Fig. 12. Spatial heterogeneity of Conditional Average Treatment Effects identifying high intervention zones.

5.4 Contextualizing causal heterogeneity: Visual evidence from representative streetscapes

To provide intuitive verification of the spatial heterogeneity revealed in Section 5.3, this section contrasts two pairs of representative case studies based on street view imagery, as shown in Fig. 13. These cases contextualize the quantitative findings by illustrating the distinct visual and socioeconomic mechanisms driving safety outcomes in different urban zones.

The first pair represents the dense urban core, where our model identified the strongest protective effects. Fig. 13 (a) depicts a high-crash site dominated by grey infrastructure, characterized by visual clutter and a lack of vertical separation. In sharp contrast, the low-crash counterpart in the same zone (Fig. 13 (b)) features continuous greenery that forms an effective visual buffer. This phenomenon may stem from the fact that in highly visually cluttered urban core environments, roadside greenery simplifies the drivers' visual field through screening effects, reducing distractions unrelated to traffic and thereby improving driving safety (Anciaes, 2023). This suggests that green elements with buffering and guiding functions should be incorporated into the design and renovation of high-density traffic environments to enhance the resilience of the road system.

The second pair illustrate the suburban region, where the protective effect of greenery was found to be weaker. While Fig. 13 (c) shows a typical low crash segment where greenery dominates the visual environment, Fig. 13 (d) reveals that high crash clusters can still occur in these settings despite high greenery coverage. The results indicate that the protective effects of greenery in this area are relatively weak. This may be attributed to the already abundant greenery, which contributes little additional visual simplification or spatial buffering, and to the lower visual stress experienced by drivers in suburban settings (Kasha et al., 2025). This suggests that in low density areas with well-established green infrastructure, further increases in greenery may not result in proportional safety benefits. Greater attention should be given to the configuration and placement of greenery, as well as the potential role of other visual environment features beyond greenery. A synthesis of these four cases indicates that the impact of greenery on traffic safety depends not only on its quantity but also on the surrounding socioeconomic context and infrastructure conditions. This highlights the necessity of adopting differentiated design and management strategies.

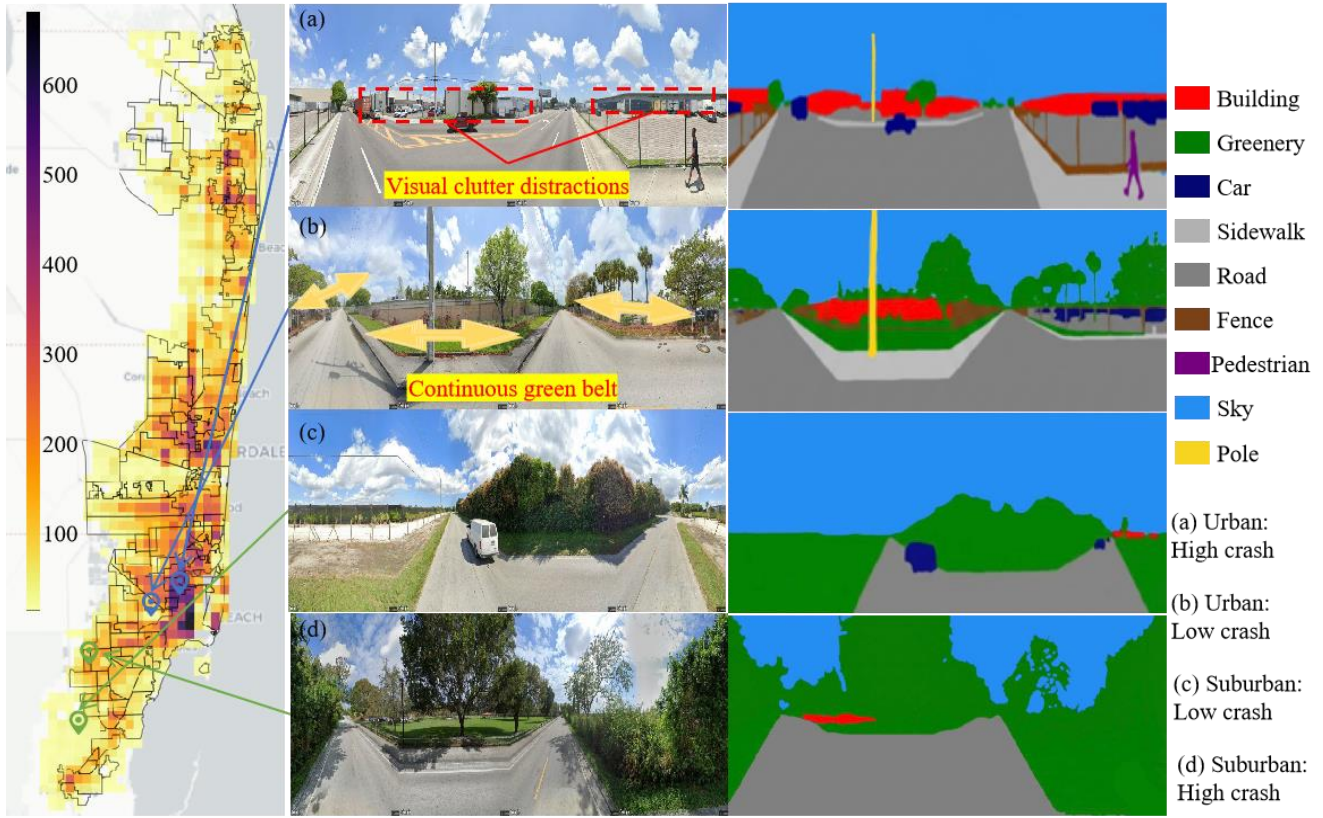

Fig. 13. Illustrative street level case studies linking crash hotspots to semantic streetscape characteristics across urban and suburban areas.

## 6. Conclusions and policy implications

Urban traffic safety remains a critical global challenge (Choudhary et al., 2024). With advances in technology, it has become feasible to leverage high resolution street view imagery to investigate the influence of the visual environment on traffic crash frequency. Existing studies have established associations between visual environment features and traffic crashes, yet most prior analyses rely on correlational evidence, leaving the causal effects of the visual environment on traffic crashes insufficiently understood. This study addresses this gap by employing high resolution street view imagery and a two-stage causal inference framework to systematically examine the causal relationship between the visual environment features and traffic crash frequency. By integrating high-resolution street view imagery with a two-stage causal inference framework—combining Double Machine Learning (DML) and Causal Forests—we systematically examined the causal mechanisms of street-level features across Southeast Florida. The methodological approach and empirical findings offer significant contributions to the fields of urban planning and transportation safety. The main findings of this study are summarized as follows:

(1) This study establishes a causal framework integrating DML and Causal Forests. Results indicate that among numerous visual features, greenery proportion emerges as the sole significant global causal determinant (ATE: −6.38). This finding validates the efficacy of the proposed framework in isolating causal signals from high-dimensional observational data, confirming that street-level greenery is a fundamental determinant of traffic safety.

(2) The Causal Forest analysis reveals that the protective effects of greenery are characterized by marked spatial and contextual heterogeneity. The safety benefits are not uniformly distributed but are significantly amplified in high-density urban cores and socioeconomically disadvantaged communities. This evidence suggests that strategic investment in green infrastructure can serve as a vital instrument for advancing both environmental quality and safety equity in vulnerable neighborhoods.

(3) The protective effects of greenery are broadly present across crash types, being particularly significant for preventing rear-end, angle, and injury crashes. Although the impact on pedestrian, bicycle, and severe injury crashes is relatively weaker, it still yields high marginal returns in areas with scarce greenery. This implies that addressing high traffic crash frequency requires shifting from mere quantitative expansion to more precisely targeted intervention strategies.

In summary, this study provides robust causal evidence that urban greenery significantly reduces traffic crash frequency, underscoring the necessity of integrating street level visual environment features into causal safety assessments. Based on these findings, we suggest that safety oriented greening policies must be meticulously tailored to local contexts and bolstered by consistent monitoring and evaluation.

Nonetheless, several limitations should be acknowledged. First, the analysis is conducted within the urban context of Southeast Florida, which may constrain the transferability of the findings to regions with different cultural or urban form characteristics. Future research should replicate the proposed framework across diverse geographic settings to assess external validity. Second, greenery is represented as an aggregate proportion, and different forms of vegetation may operate through distinct behavioral and perceptual mechanisms. Future studies should disaggregate green space into finer typologies using advanced image segmentation techniques to refine causal interpretation. Finally, the reliance on cross sectional data limits the ability to capture dynamic causal processes. Future work should incorporate longitudinal or seasonal street view imagery to examine how temporal changes in greenery influence traffic safety outcomes over time.

**CRediT authorship contribution statement**

**Lishan Sun**: Formal analysis, Validation, Supervision, Writing - review & editing. **Yujia Cheng**: Conceptualization, Data curation, Formal analysis, Writing – original draft, Writing – review & editing, Methodology. **Pengfei Cui**: Formal analysis, Writing – original draft, Writing – review & editing, Investigation, Validation. **Lei Han**: Writing – review & editing, Data curation, Formal analysis, Validation. **Mohamed Abdel-Aty**: Writing – review & editing, Supervision. **Yunhan Zheng**: Writing – review & editing, Formal analysis, Validation. **Xingchen Zhang**: Writing – review & editing, Supervision.

**Declaration of competing interest**

The authors declare that they have no known competing financial interests or personal relationships that could have appeared to influence the work reported in this paper.

**Acknowledgments**

This work was supported by National Natural Science Foundation of China (No. 52472317), Beijing Natural Science Foundation (No. L231023) and Fundamental Research Funds for Beijing Municipal Universities.

# References

Abécassis, J., Dumas, É., Alberge, J., Varoquaux, G., 2025. From prediction to prescription: Machine learning and causal inference for the heterogeneous treatment effect. Annu. Rev. Biomed. Data Sci. 8 1 , 381–404. doi:10.1146/annurev-biodatasci-103123-095750

Alisan, O., Ozguven, E.E., 2024. An analysis of the spatial variations in the relationship between built environment and severe crashes. Isprs Int. J. Geo-inf. 13 12 , 465. doi:10.3390/ijgi13120465

Anciaes, P., 2023. Effects of the roadside visual environment on driver wellbeing and behaviour – a systematic review. Transport Rev. 43 4 , 571–598. doi:10.1080/01441647.2022.2133189

Asadi Ghalehni, S., Afghari, A.P., 2026. A hybrid statistical-machine learning methodology for addressing endogeneity and temporal instability in speeding-crash frequency relationships. Accid. Anal. Prev. 224, 108284. doi:10.1016/j.aap.2025.108284

Asadi, M., Ulak, M.B., Geurs, K.T., Weijermars, W., Schepers, P., 2022. A comprehensive analysis of the relationships between the built environment and traffic safety in the dutch urban areas. Accid. Anal. Prev. 172, 106683. doi:10.1016/j.aap.2022.106683

Athey, S., Tibshirani, J., Wager, S., 2019. Generalized random forests. Ann. Stat. 47 2 . doi:10.1214/18-AOS1709

Bach, P., Kurz, M.S., Chernozhukov, V., Spindler, M., Klaassen, S., 2024. **DoubleML** : An object-oriented implementation of double machine learning in *R*. J. Stat. Softw. 108 3 , 0–56. doi:10.18637/jss.v108.i03

Cai, Q., Abdel-Aty, M., Zheng, O., Wu, Y., 2022. Applying machine learning and google street view to explore effects of drivers' visual environment on traffic safety. Transportation Research Part C: Emerging Technologies 135, 103541. doi:10.1016/j.trc.2021.103541

Chen, L.-W., Yavuz, I., Cheng, Y., Wahed, A.S., 2020. Cumulative incidence regression for dynamic treatment regimens. Biostatistics 21 2 , e113–e130. doi:10.1093/biostatistics/kxy062

Cheng, D., Li, J., Liu, L., Le, T.D., Liu, J., Yu, K., 2022. Sufficient dimension reduction for average causal effect estimation. Data Min. Knowl. Discovery 36 3 , 1174–1196. doi:10.1007/s10618-022-00832-5

Chernozhukov, V., Chetverikov, D., Demirer, M., Duflo, E., Hansen, C., Newey, W., Robins, J., 2018. Double/debiased machine learning for treatment and structural parameters. Econom. J. 21 1 , C1–C68. doi:10.1111/ectj.12097

Choi, D., Ewing, R., 2021. Effect of street network design on traffic congestion and traffic safety. J. Transp. Geogr. 96, 103200. doi:10.1016/j.jtrangeo.2021.103200

Choudhary, A., Garg, R.D., Jain, S.S., Khan, A.B., 2024. Impact of traffic and road infrastructural design variables on road user safety – a systematic literature review. Int. J. Crashworthiness 29 4 , 583–596. doi:10.1080/13588265.2023.2274641

Credit, K., Lehnert, M., 2024. A structured comparison of causal machine learning methods to assess heterogeneous treatment effects in spatial data. J. Geogr. Syst. 26 4 , 483–510. doi:10.1007/s10109-023-00413-0

Cui, P., Abdel-Aty, M., Han, L., Yang, X., 2025. Multiscale geographical random forest: A novel spatial ML approach for traffic safety modeling integrating street-view semantic visual features. Transportation Research Part C: Emerging Technologies 179, 105299. doi:10.1016/j.trc.2025.105299

Cutler, A., Cutler, D.R., Stevens, J.R., 2012. Random forests, in: Zhang, C., Ma, Y. (Eds.), Ensemble Machine Learning. Springer New York, New York, NY, pp. 157–175. doi:10.1007/978-1-4419-9326-7_5

Ehsani, J.P., Michael, J.P., MacKENZIE, E.J., 2023. The future of road safety: Challenges and opportunities. Milbank Q. 101 S1 , 613–636. doi:10.1111/1468-0009.12644

Fan, Z., Feng, C.-C., Biljecki, F., 2025. Coverage and bias of street view imagery in mapping the urban environment. Comput. Environ. Urban Syst. 117, 102253. doi:10.1016/j.compenvurbsys.2025.102253

Fan, Z., Zhang, F., Loo, B.P.Y., Ratti, C., 2023. Urban visual intelligence: Uncovering hidden city profiles with street view

images. Proc. Natl. Acad. Sci. U.S.A. 120 27 , e2220417120. doi:10.1073/pnas.2220417120

Fang, Y.-N., Tian, J., Namaiti, A., Zhang, S., Zeng, J., Zhu, X., 2024. Visual aesthetic quality assessment of the streetscape from the perspective of landscape-perception coupling. Environ. Impact Assess. Rev. 106, 107535. doi:10.1016/j.eiar.2024.107535

Feuerriegel, S., Frauen, D., Melnychuk, V., Schweisthal, J., Hess, K., Curth, A., Bauer, S., Kilbertus, N., Kohane, I.S., Van Der Schaar, M., 2024. Causal machine learning for predicting treatment outcomes. Nat. Med. 30 4 , 958–968. doi:10.1038/s41591-024-02902-1

Fortuin, V., 2022. Priors in bayesian deep learning: A review. Int. Stat. Rev. 90 3 , 563–591. doi:10.1111/insr.12502

Friedman, J.H., 2001. Greedy function approximation: A gradient boosting machine. Ann. Stat. 29 5 . doi:10.1214/aos/1013203451

Gan, J., Su, Q., Li, Linheng, Ju, Y., Li, Linchao, 2025. Urban traffic accident frequency modeling: An improved spatial matrix construction method. J. Adv. Transp. 2025 1 , 1923889. doi:10.1155/atr/1923889

Gerike, R., Koszowski, C., Schröter, B., Buehler, R., Schepers, P., Weber, J., Wittwer, R., Jones, P., 2021. Built environment determinants of pedestrian activities and their consideration in urban street design. Sustainability 13 16 , 9362. doi:10.3390/su13169362

Graham, D.J., 2025. Causal inference for transport research. Transp. Res. Part A Policy Pract. 192, 104324. doi:10.1016/j.tra.2024.104324

Guan, F., Fang, Z., Wang, L., Zhang, X., Zhong, H., Huang, H., 2022. Modelling people's perceived scene complexity of real-world environments using street-view panoramas and open geodata. Isprs J. Photogramm. Remote Sens. 186, 315–331. doi:10.1016/j.isprsjprs.2022.02.012

Guo, Y., Li, M., Li, K., Li, H., Li, Y., 2024. Unraveling the determinants of traffic incident duration: A causal investigation using the framework of causal forests with debiased machine learning. Accid. Anal. Prev. 208, 107806. doi:10.1016/j.aap.2024.107806

Hamim, O.F., Ukkusuri, S.V., 2024. Towards safer streets: A framework for unveiling pedestrians' perceived road safety using street view imagery. Accid. Anal. Prev. 195, 107400. doi:10.1016/j.aap.2023.107400

Hirano, K., Imbens, G.W., 2004. The propensity score with continuous treatments, in: Gelman, A., Meng, X. (Eds.), Wiley Series in Probability and Statistics. Wiley, pp. 73–84. doi:10.1002/0470090456.ch7

Høye, A.K., Hesjevoll, I.S., 2020. Traffic volume and crashes and how crash and road characteristics affect their relationship – a meta-analysis. Accid. Anal. Prev. 145, 105668. doi:10.1016/j.aap.2020.105668

Imbens, G.W., Wooldridge, J.M., 2009. Recent developments in the econometrics of program evaluation. J. Econ. Lit. 47 1 , 5–86. doi:10.1257/jel.47.1.5

Ito, K., Bansal, P., Biljecki, F., 2024. Examining the causal impacts of the built environment on cycling activities using time-series street view imagery. Transp. Res. Part A Policy Pract. 190, 104286. doi:10.1016/j.tra.2024.104286

Jia, L., Huang, C., Du, N., 2024. Drivers' situational awareness of surrounding vehicles during takeovers: Evidence from a driving simulator study. Transp. Res. Part F Psychol. Behav. 106, 340–355. doi:10.1016/j.trf.2024.08.016

Kasha, A., Tefft, B.C., Steinbach, R., 2025. Community vulnerability influences traffic safety differently in urban, suburban, and rural areas. J. Transp. Health 44, 102146. doi:10.1016/j.jth.2025.102146

Ke, G., Meng, Q., Finley, T., Wang, T., Chen, W., Ma, W., Ye, Q., Liu, T.-Y., n.d. LightGBM: A highly efficient gradient boosting decision tree 0–9.

Kim, S.H., 2023. How heterogeneity has been examined in transportation safety analysis: A review of latent class modeling applications. Anal. Methods Accid. Res. 40, 100292. doi:10.1016/j.amar.2023.100292

Komura, T., Bargagli-Stoffi, F.J., Shiba, K., Inoue, K., 2025. Two-step pragmatic subgroup discovery for heterogeneous

treatment effects analyses: Perspectives toward enhanced interpretability. Eur. J. Epidemiol. 40 2 , 141–150. doi:10.1007/s10654-025-01215-y

Larkin, A., Gu, X., Chen, L., Hystad, P., 2021. Predicting perceptions of the built environment using GIS, satellite and street view image approaches. Landscape Urban Plann. 216, 104257. doi:10.1016/j.landurbplan.2021.104257

Lechner, M., Mareckova, J., 2024. Comprehensive causal machine learning. doi:10.48550/ARXIV.2405.10198

Lee, S., Ma, Y., De Luna, X., 2025. Covariate balancing for causal inference on categorical and continuous treatments. Econom. Stat. 33, 304–329. doi:10.1016/j.ecosta.2022.01.007

Li, S., Pu, Z., Cui, Z., Lee, S., Guo, X., Ngoduy, D., 2024. Inferring heterogeneous treatment effects of crashes on highway traffic: A doubly robust causal machine learning approach. Transp. Res. Part C Emerging Technol. 160, 104537. doi:10.1016/j.trc.2024.104537

Li, X., Yu, S., Huang, X., Dadashova, B., Cui, W., Zhang, Z., 2022. Do underserved and socially vulnerable communities observe more crashes? A spatial examination of social vulnerability and crash risks in texas. Accid. Anal. Prev. 173, 106721. doi:10.1016/j.aap.2022.106721

Li, X., Zhang, C., Li, W., Ricard, R., Meng, Q., Zhang, W., 2015. Assessing street-level urban greenery using google street view and a modified green view index. Urban For. Urban Greening 14 3 , 675–685. doi:10.1016/j.ufug.2015.06.006

Li, Y., Peng, L., Wu, C., Zhang, J., 2022. Street view imagery (SVI) in the built environment: A theoretical and systematic review. Buildings 12 8 , 1167. doi:10.3390/buildings12081167

Li, Y., Zhang, Yecheng, Wu, Q., Xue, R., Wang, X., Si, M., Zhang, Yuyang, 2023. Greening the concrete jungle: Unveiling the co-mitigation of greenspace configuration on PM2.5 and land surface temperature with explanatory machine learning. Urban For. Urban Greening 88, 128086. doi:10.1016/j.ufug.2023.128086

Liu, Y., Chen, T., Chung, H., Jang, K., Xu, P., 2025. Is there an emotional dimension to road safety? A spatial analysis for traffic crashes considering streetscape perception and built environment. Anal. Methods Accid. Res. 46, 100374. doi:10.1016/j.amar.2025.100374

Liu, Y., Gu, X., Wang, Z., Anderson, A., 2024. Urban greenery distribution and its link to social vulnerability. Urban For. Urban Greening 101, 128542. doi:10.1016/j.ufug.2024.128542

Lundberg, S.M., Lee, S.-I., n.d. A unified approach to interpreting model predictions 0–10.

Metz-Peeters, M., 2025. Mandatory speed limits and crash frequency on motorways — a causal machine learning approach. Transp. Res. Part A Policy Pract. 200, 104616. doi:10.1016/j.tra.2025.104616

Mohammed, S., Alkhereibi, A.H., Abulibdeh, A., Jawarneh, R.N., Balakrishnan, P., 2023. GIS-based spatiotemporal analysis for road traffic crashes; in support of sustainable transportation planning. Transp. Res. Interdiscip. Perspect. 20, 100836. doi:10.1016/j.trip.2023.100836

Naseralavi, S., 2025. Exploring the impact of driver sex, driver age, area type, and lighting conditions on rear-end collision severity. COMPUT. RES. PROG. IN APPL. SCI. ENG. 1–17. doi:10.82042/crpase.11.3.2957

Nicholls, V.I., Wiener, J., Meso, A.I., Miellet, S., 2024. The impact of perceptual complexity on road crossing decisions in younger and older adults. Sci. Rep. 14 1 , 479. doi:10.1038/s41598-023-49456-9

Park, J., Lee, S., 2026. Do visual attributes of streetscapes affect car crashes? Applications of computer vision techniques and machine learning. Travel Behav. Soc. 42, 101153. doi:10.1016/j.tbs.2025.101153

Pi, M., Yeon, H., Son, H., Jang, Y., 2021. Visual cause analytics for traffic congestion. IEEE Trans. Vis. Comput. Graphics 27 3 , 2186–2201. doi:10.1109/TVCG.2019.2940580

Shen, K., Liu, J., Liu, X., 2025. Understanding the impact of street environments on traffic crash risk from the perspective of aging people: An interpretable machine learning approach. Isprs Int. J. Geo-inf. 14 7 , 248. doi:10.3390/ijgi14070248

Snoek, J., Larochelle, H., Adams, R.P., n.d. Practical bayesian optimization of machine learning algorithms 0–9.

Souto, H.G., Neto, F.L., 2024. Advancing causal inference: a nonparametric approach to ATE and CATE estimation with continuous treatments. doi:10.48550/ARXIV.2409.06593

Strudel, R., Garcia, R., Laptev, I., Schmid, C., 2021. Segmenter: Transformer for semantic segmentation, in: 2021 IEEE/CVF International Conference on Computer Vision (ICCV). Presented at the 2021 IEEE/CVF International Conference on Computer Vision (ICCV), IEEE, Montreal, QC, Canada, pp. 7242–7252. doi:10.1109/ICCV48922.2021.00717

Sung, H., Lee, S., Cheon, S., Yoon, J., 2022. Pedestrian safety in compact and mixed-use urban environments: Evaluation of 5D measures on pedestrian crashes. Sustainability 14 2 , 646. doi:10.3390/su14020646

Van Vogt, E., Gordon, A.C., Diaz-Ordaz, K., Cro, S., 2025. Application of causal forests to randomised controlled trial data to identify heterogeneous treatment effects: A case study. BMC Med. Res. Methodol. 25 1 , 50. doi:10.1186/s12874-025-02489-2

Wager, S., Athey, S., 2018. Estimation and inference of heterogeneous treatment effects using random forests. J. Am. Stat. Assoc. 113 523 , 1228–1242. doi:10.1080/01621459.2017.1319839

Wang, C., Abdel-Aty, M., Zhai, S., Uddin, A.S.M.N., Islam, Z., 2026. From prediction to explanation: A machine learning and causal mediation framework for roadway crash risk with connected vehicle data. Transportation Research Part C: Emerging Technologies 183, 105479. doi:10.1016/j.trc.2025.105479

Wang, Y., Jiao, Y., Fu, L., Shangguan, Q., 2025. Exploring causal factor in highway–railroad-grade crossing crashes: A comparative analysis. Infrastructures 10 8 , 216. doi:10.3390/infrastructures10080216

White, E.O., Meixler, M.S., 2024. Assessing large-scale roadside tree removal using aerial imagery and crash analysis: A difference-in-differences approach. Landscape Urban Plann. 244, 104980. doi:10.1016/j.landurbplan.2023.104980

Xu, C., Zhang, Z., Fu, F., Yao, W., Su, H., Hu, Y., Rong, D., Jin, S., 2023. Analysis of spatiotemporal factors affecting traffic safety based on multisource data fusion. J. Transp. Eng. A: Syst. 149 10 , 4023098. doi:10.1061/JTEPBS.TEENG-7990

Yang, S., Zhou, L., Zhang, Z., Li, H., Guo, L., Sun, X., Song, T., 2025. Revisiting the causal relationship between the built environment, automobile ownership, and mode choice using double machine learning. J. Transp. Geogr. 128, 104379. doi:10.1016/j.jtrangeo.2025.104379

Ye, X., Li, S., Gong, W., Li, Xiao, Li, Xinyu, Dadashova, B., Li, W., Du, J., Wu, D., 2025. Street view imagery in traffic crash and road safety analysis: a review. Appl. Spatial Anal. Policy 18 2 , 50. doi:10.1007/s12061-025-09653-7

Ye, Y., Zhong, C., Suel, E., 2024. Unpacking the perceived cycling safety of road environment using street view imagery and cycle accident data. Accid. Anal. Prev. 205, 107677. doi:10.1016/j.aap.2024.107677

Yue, H., 2025. Investigating streetscape environmental characteristics associated with road traffic crashes using street view imagery and computer vision. Accid. Anal. Prev. 210, 107851. doi:10.1016/j.aap.2024.107851

Zaidi, S.Z., Wang, X., Azati, Y., Li, J., Fan, T., Quddus, M., 2025. Heterogeneous and differential treatment effect analysis of safety improvements on freeways using causal inference. Accid. Anal. Prev. 220, 108173. doi:10.1016/j.aap.2025.108173

Zhang, S., Sze, N.N., Abdel-Aty, M., 2025. What street view imagery features favour driving? A copula model for driver distraction and driving performance. Travel Behav. Soc. 41, 101068. doi:10.1016/j.tbs.2025.101068

Zhang, Y., Li, H., Sze, N.N., Ren, G., 2021. Propensity score methods for road safety evaluation: Practical suggestions from a simulation study. Accid. Anal. Prev. 158, 106200. doi:10.1016/j.aap.2021.106200

Zhu, M., Sze, N.N., Newnam, S., 2022. Effect of urban street trees on pedestrian safety: A micro-level pedestrian casualty model using multivariate bayesian spatial approach. Accid. Anal. Prev. 176, 106818. doi:10.1016/j.aap.2022.106818